\documentclass[10pt,journal,final]{IEEEtran}
\IEEEoverridecommandlockouts
\usepackage{setspace}
\usepackage{cite}
\usepackage{amsmath,amssymb,amsfonts}
\usepackage{makecell}
\usepackage{algorithmic}
\usepackage{graphicx}
\usepackage{textcomp}
\usepackage{hyperref}
\usepackage{pifont}
\usepackage{xcolor}
\usepackage[switch]{lineno}
\usepackage{colortbl}
\usepackage{titlesec}
\def\BibTeX{{\rm B\kern-.05em{\sc i\kern-.025em b}\kern-.08em
    T\kern-.1667em\lower.7ex\hbox{E}\kern-.125emX}}
\begin{document}

\definecolor{yyellow}{RGB}{235,211,70}
\definecolor{rred}{RGB}{245, 152, 153}
\definecolor{bblue}{RGB}{80, 181, 255}

\title{Is Underwater Image Enhancement \\ All Object Detectors Need?}

\author{Yudong~Wang, Jichang~Guo, Wanru~He, Huan~Gao, Huihui~Yue, Zenan~Zhang, and Chongyi~Li\\
\thanks{Y. Wang and J. Guo are with the School of Electrical and Information Engineering, Tianjin University, Tianjin 300072, China, and also with Shanghai Artificial Intelligence Laboratory, Shanghai 200232, China (e-mail: yudongwang@tju.edu.cn; jcguo@tju.edu.cn).}
\thanks{W. He, H. Gao, H. Yue, and Z. Zhang are with the School of Electrical and Information Engineering, Tianjin University, Tianjin 300072, China (e-mail: hewan\_ru@tju.edu.cn; gh99@tju.edu.cn; yuehuihui@tju.edu.cn; znzhang@tju.edu.cn;)}
\thanks{C. Li is with the School of Computer Science, Nankai University, Tianjin 300071, China  (e-mail: lichongyi@nankai.edu.cn).}
\thanks{J. Guo is the corresponding author.}
\thanks{This work was supported in part by the National Natural Science Foundation of China (No.62171315), in part by the National Key Research and Development Program of China (No. 2022ZD0160400), and in part by Tianjin Research Innovation Project for Postgraduate Students (No.2021YJSB153).}
}

\maketitle
\vspace{-2cm
}
\begin{abstract}
Underwater object detection is a crucial and challenging problem in marine engineering and aquatic robot. 
The difficulty is partly because of the degradation of underwater images caused by light selective absorption and scattering. 
Intuitively, enhancing underwater images can benefit high-level applications like underwater object detection. 
However, it is still unclear whether all object detectors need underwater image enhancement as pre-processing.
We therefore pose the questions \textit{“Does underwater image enhancement really improve underwater object detection?”} and \textit{“How does underwater image enhancement contribute to underwater object detection?”}.
With these two questions, we conduct extensive studies.
Specifically, we use 18 state-of-the-art underwater image enhancement algorithms, covering traditional, CNN-based, and GAN-based algorithms, to pre-process underwater object detection data. Then, we retrain 7 popular deep learning-based object detectors using the corresponding results enhanced by different algorithms, obtaining 126 underwater object detection models. 
Coupled with 7 object detection models retrained using raw underwater images, we employ these 133 models to comprehensively analyze the effect of underwater image enhancement on underwater object detection. 
We expect this study can provide sufficient exploration to answer the aforementioned questions and draw more attention of the community to the joint problem of underwater image enhancement and underwater object detection. 
The pre-trained models and results are publicly available and will be regularly updated. 
Project page: \url{https://github.com/BIGWangYuDong/lqit/tree/main/configs/detection/uw_enhancement_affect_detection}.

\end{abstract}


\section{Introduction}

In recent years, underwater object detection has become a crucial topic as it plays a vital role in several applications, such as the repair and maintenance of sub-aquatic structures, marine environmental protection, and marine engineering \cite{anwar2020diving}. However, detecting underwater objects is challenging. The biggest obstacle is that raw underwater images usually have low quality such as low contrast and brightness, color deviations, blurry details, uneven bright specks, \textit{etc}. These degradation issues are mainly caused by backscattering, light selective absorption, and scattering in water.

\begin{figure}[tbp]
  \centering
  \includegraphics[width=1\linewidth]{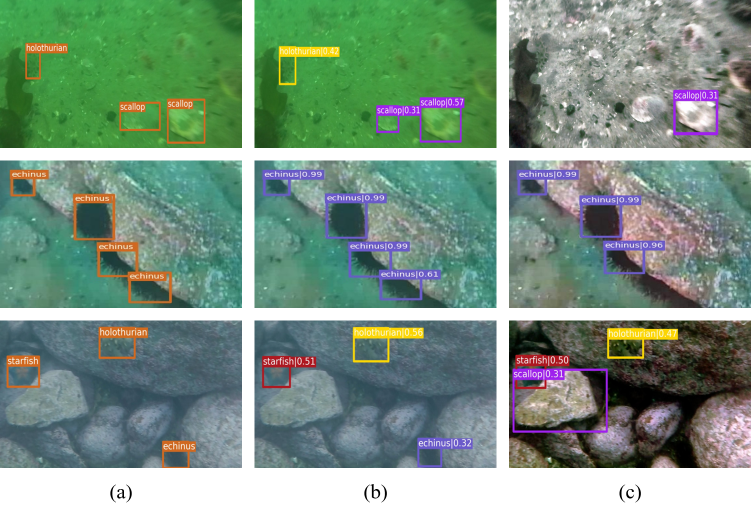}
  \caption{\textbf{Samples of real-world underwater image enhancement and object detection.} (a) The ground truth of underwater object detection. (b) The detection results of ATSS \cite{zhang2020bridging} retrained on raw underwater images. (c) The detection results of ATSS \cite{zhang2020bridging} retrained on the underwater images pre-processed by ACDC \cite{zhang2022underwater}, UIEC\^{}2Net \cite{wang2021uiec}, and DMIL-HDP \cite{li2016mlhp} (from top to bottom). (Zoom in for best view)}
  \label{fig1}
\end{figure}

Intuitively, underwater image enhancement is a direct way to improve the performance of underwater object detectors as it may recover more information from underwater images.
It has even become common sense that improving the visual quality of underwater images can improve the performance of high-level tasks.
However, a comprehensive study on this issue is still scarce.
Chen \textit{et al.} \cite{chen2020reveal} have briefly discussed the impact of underwater image enhancement on object detection and drawn a preliminary conclusion: image enhancement will inhibit underwater object detection performance. 
However, due to the insufficient experiments and analysis in Chen \textit{et al.}'s work \cite{chen2020reveal}, it is too early to answer the questions \textit{“Does underwater image enhancement improve object detection?”} and \textit{“How does underwater image enhancement contribute to object detection?”}.

In this paper, we try to answer these two questions by jointly analyzing underwater image enhancement and underwater object detection. 
To be specific, we first select 18 underwater image enhancement algorithms, including traditional physical model-free, physical model-based algorithms, and deep learning-based CNN and GAN algorithms to process the underwater object detection data.
Then, qualitative and quantitative comparisons are conducted to analyze the results enhanced by different underwater image enhancement algorithms.
Furthermore, the data pre-processed by different underwater enhancement algorithms (we define each enhanced dataset as a data domain) are used to retrain 7 popular deep learning-based object detectors, including two-stage and one-stage detectors with anchor-free or anchor-based methods. 
We show some samples of enhancement and detection results in Fig. \ref{fig1}. In comparison to the results of ATSS \cite{zhang2020bridging} retrained on the raw underwater images, the same detector retrained on the underwater images pre-processed by ACDC \cite{zhang2022underwater}, UIEC\^{}2Net \cite{wang2021uiec}, and DMIL-HDP \cite{li2016mlhp} detects the background as foreground or cannot fully detect the accurate object.
With the enhancement and detection results, quantitative and qualitative comparisons under diverse evaluation metrics, including AP, TIDE \cite{tideeccv2020}, Precision-Recall curves, and Class Activation Map (CAM) visualization are adopted to analyze the effect of underwater image enhancement on underwater object detection.
Finally, we discuss and provide an outlook on the future research directions of underwater image enhancement, underwater image quality assessment metrics, and underwater object detection.

\textbf{The main contributions of this paper are summarized as three-fold.} 
\begin{itemize}
\item We conduct the first comprehensive empirical study on the impact of underwater image enhancement on object detection. 
\item  Through extensive experiments and analysis, we unveil the shortcomings of existing underwater image enhancement algorithms, including the limited robustness and adaptability, especially for underwater object detection.
\item  We analyze the limitations of the existing underwater image quality assessment metrics, which not only have a gap with human visual perception but also cannot directly represent the performance of subsequent high-level tasks (\textit{e.g.} object detection). 
\end{itemize}

Note that the goal of this paper is not to propose a new method to improve the performance of underwater object detection or enhance the quality of underwater images. 
Instead, we study whether the performance of underwater object detection can be improved by pre-processing training and testing data and in what ways the enhanced images affect object detection. 
We expect that this work can attract more community interest to further study joint object detection and image enhancement, such as developing new joint algorithms to improve both object detection and image visual quality.

\section{Relate Work}

\noindent
\textbf{Object Detection.} 
Current CNN-based object detection can be simply divided into two-stage and one-stage detectors. 
Ren \textit{et al.} \cite{ren2015faster} proposed Faster R-CNN, as a classic detector, which establishes the dominance of two-stage detectors.
Faster R-CNN consists of a separate region proposal network (RPN) and a region-wise prediction network (R-CNN) \cite{girshick2014rich, girshick2015fast} to detect objects. 
Inspired by Faster R-CNN, many R-CNN-based two-stage detectors were proposed, including architecture improvements \cite{cai2018cascade}, more efficient training strategies \cite{najibi2019autofocus, singh2018analysis}, and more useful loss functions \cite{he2019bounding, wang2017fast}.
The one-stage detector also obtains attention because of its high computational efficiency. 
As a pioneer work, Liu \textit{et al.} \cite{liu2016ssd} proposed the SSD for real-time detection. 
Thereafter, plenty of works were proposed to improve SSD's performance in different aspects, including assign strategies \cite{zhang2020bridging, zhang2021learning}, new loss function \cite{lin2017focal, chen2019towards}, and architecture improvements \cite{liu2018receptive, nie2019enriched}. 
Besides, object detectors can be divided into anchor-based and anchor-free methods. The former is based on default anchors to adjust the outputs, such as Faster R-CNN \cite{ren2015faster}, SSD \cite{liu2016ssd}, and RetinaNet \cite{lin2017focal}, while the latter does not generate anchors, such as FCOS \cite{tian2019fcos}.

\noindent
\textbf{Underwater Image Enhancement.} 
Underwater image enhancement can be roughly classified into two groups: traditional and deep learning-based methods. 
The former can be further divided into physical free-based and physical model-based methods. 
Physical free-based methods mainly adjust the pixel values for visual quality improvement \cite{kang2022perception,zhuang2021bayesian,liang2022gifm, zhang2022underwater2,ancuti2012enhancing,ancuti2019color}. 
However, due to the omission of an underwater imaging mechanism, physical free-based methods tend to underperform when encountering diverse and challenging underwater scenes. 
In contrast, physical model-based underwater image enhancement methods are mainly based on prior information, such as red channel prior \cite{galdran2015automatic}, underwater dark channel prior \cite{drews2016underwater}, and general dark channel prior \cite{peng2018generalization}. 
The success of the deep learning strategy inspires its application in degraded image enhancement, such as dehazing\cite{guo2022image,wu2023ridcp}, image colorization \cite{anwar2020image}, low-light image enhancement \cite{liang2023iterative,dai2023nighttime,li2023embedding}, and underwater image enhancement \cite{anwar2020diving}. 
Hence, there are several attempts to improve the performance of underwater image enhancement through deep learning strategies.
In addition to simply organizing them into CNN-based and GAN-based methods, we can also divide them into five categories according to the architectural differences: 
1) encode-decode structure \cite{sun2018deep,uplavikar2019all,fabbri2018enhancing,guo2022underwater}, which is one of the structures commonly used in low-level tasks;
2) high-performance modular or block \cite{li2020underwater,guo2019underwater}, which focuses on improving the feature extraction capability of the network;
3) multiple branches \cite{wang2017deep, li2019underwater,li2019fusion, wang2021uiec}, which aim to learn different features and fuses together;
4) multiple generators \cite{li2018emerging, lu2019multi, ye2018underwater}, which enhance underwater images by sharing the features of generators; and
5) deep learning methods combined with physical models \cite{hou2018joint, cao2018underwater,li2017watergan, li2021underwater}, improving the performance of enhancement and restoration by predicting the depth map or transmission map of the underwater image. 
Although these methods have significantly improved the underwater image quality, they do not verify whether the enhancement can improve the performance of underwater object detection.

\noindent
\textbf{Joint Image Enhancement and High-level Applications.}
In reality, the acquired images may contain some degradation issues, thus affecting high-level applications. 
Intuitively, enhancing such degraded images can benefit high-level applications.
However, several studies on the relationship between image enhancement and high-level applications show that the enhancement cannot improve the performance of high-level tasks.
For example, Pei \textit{et al.} \cite{pei2018does, pei2019effects} found that the existing image dehazing and deblurring methods cannot improve or even reduce the image classification performance. 
\cite{sakaridis2018semantic} found that dehazing cannot improve the semantic segmentation performance, and \cite{li2019deraining, hnewa2020object} suggested that the existing de-raining algorithms deteriorate the detection performance compared to directly using the rainy image. 
\cite{xiao2020making, dai2018dark, guo2020zero} drew a similar conclusion in low-light object detection and semantic segmentation.
For underwater scenes, Zhuang \textit{et al.} \cite{zhuang2022underwater} applied underwater image enhancement results in segmentation, saliency detection, keypoint detection, and depth estimation, and the experiment results show that underwater image enhancement will improve the performance of these applications. 
However, these application experiments are only concluded on several cherry-picked cases.
Furthermore, \cite{chen2020reveal, liu2020real} concluded that most of underwater image enhancement methods would inhibit the object detection performance, and the experiments in \cite{liu2022novel, liu2022twin} demonstrate this finding.
These studies generally report that most of image enhancement methods cannot benefit high-level applications but do not fully analyze the inherent reasons. 
Moreover, some studies \cite{liu2022twin, jiang2021underwater} trying to combine underwater image enhancement and object detection methods to simultaneously improve the performance of underwater image enhancement and object detection in an end-to-end way. 
Although the combined method can improve the performance, they do not answer the question: \textit{“Does underwater image enhancement improve object detection?”} and \textit{“How does underwater image enhancement contribute to object detection?”}.
In this paper, we study whether the performance of underwater object detection can be improved by pre-processing  training and testing data and in what ways the enhanced images affect object detection via more comprehensive and sufficient experiments.

\section{Investigating the Effect of Underwater Image Enhancement on Object Detection}
In this section, we investigate whether and how underwater image enhancement contributes to object detection. 
First, we describe the experiment settings. 
Second, the pre-processing results on underwater object detection are analyzed qualitatively and quantitatively.
Third, the experimental results on different object detectors retrained using the results enhanced by different underwater image enhancement algorithms are reported and also analyzed quantitatively and qualitatively.
Especially, the TIDE \cite{tideeccv2020}, a general toolbox that computes and evaluates the effect of object detection on overall performance, and the visualization of feature maps are used to analyze the effect of underwater image enhancement on object detection.

\subsection{Experimental Settings}\label{exp_setting}

\noindent
\textbf{Underwater Object Detection Dataset.} 
All experiments are implemented on the Underwater Robot Professional Contest dataset (URPC2020)\footnote{\url{https://www.heywhale.com/home/competition/5e535a612537a0002ca864ac/content/0}} \cite{liu2021dataset}, which contains 5,543 underwater images, covering four categories: holothurian, echinus, scallop, and starfish. 
Our study randomly divides the URPC2020 dataset into training and testing groups with 4,434 and 1,019 images, respectively.

Moreover, we also performed experiments on Real-world Underwater Object Detection (RUOD) dataset \cite{fu2023rethinking}, which is the largest underwater image dataset with object bounding boxes. 
In detail, RUOD contains 14,000 images (9,800 for training and 4,200 for testing) with more than 74,000 bounding boxes, covering ten categories: fish, echinus, corals, starfish, holothurian, scallop, diver, cuttlefish, turtle, and jellyfish.
The experimental results of Faster R-CNN \cite{ren2015faster}, Cascade R-CNN \cite{cai2018cascade}, RetinaNet \cite{lin2017focal}, FCOS \cite{tian2019fcos}, ATSS \cite{zhang2020bridging}, TOOD \cite{feng2021tood}, and SSD \cite{liu2016ssd} that training with raw and different enhanced RUOD dataset \cite{fu2023rethinking} and testing with the same domain are in Appendix \ref{appendix_A}.

\noindent
\textbf{Underwater Image Enhancement Algorithm.} 
To verify the effect of underwater image enhancement on object detection, we select 13 underwater image enhancement algorithms. 
To comprehensively discuss the advantages and disadvantages of different underwater image enhancement algorithms, we  select both traditional methods including classical physical free-based (Histogram Equalization (HE) \cite{hummel1977image}, Contrast Limited Adaptive Histogram Equalization (CLAHE) \cite{zuiderveld1994contrast}, White
Balance (WB) \cite{ebner2007color}, and ACDC \cite{zhang2022underwater}) and typical physical model-based (UDCP \cite{drews2016underwater}, DMIL-HDP \cite{li2016mlhp}, and ULAP \cite{song2018rapid}) and recent deep learning-based methods including CNN-based (UWCNN (type-3) \cite{li2020underwater}, DUIENet \cite{li2019underwater}, CHE-GLNet \cite{fu2020underwater}, UIEC\^{}2Net \cite{wang2021uiec}, UColor \cite{li2021underwater}, and SGUIE \cite{qi2022sguie}),
and GAN-based (WaterGAN \cite{li2017watergan}, UGAN \cite{fabbri2018enhancing}, TUDA \cite{wang2023domain}, TOPAL \cite{jiang2022target}, and TACL \cite{liu2022twin}).
We run the source code of these underwater image enhancement algorithms to produce the corresponding results.

\noindent
\textbf{Object Detector.} 
We select 7 deep learning-based detectors, including two-stage detectors (Faster R-CNN \cite{ren2015faster} and Cascade R-CNN \cite{cai2018cascade}) and one-stage detectors (SSD \cite{liu2016ssd}, RetinaNet \cite{lin2017focal}, FCOS \cite{tian2019fcos}, ATSS \cite{zhang2020bridging}, and TOOD \cite{feng2021tood}). 
Each detector is respectively retrained using 14 domain datasets (raw dataset and 13 enhanced datasets). 
Note that each detector's training and evaluation are based on an identical data domain. 
To avoid fluctuation, we set the seed to 0 to reduce the interference caused by randomness. 
As for training details, all detectors use the same SGD optimizer with 0.9 momentum and $5\times{10^{-4}}$ weight decay. 
As for data augmentation, except for SSD, all detectors resized the input image to have a shorter side of 800 while the longer side is kept less than 1333, and using horizontal flipping with a probability of 0.5. During testing time, we resize the input image in the same way as the training phase (\textit{i.e.}, the shorter side is resized to 800 while the longer side is kept less than 1333). 
The data augmentation of SSD \cite{liu2016ssd} is consistent with the original paper, and it should be emphasized that we resize the images to 300 during training and testing time.
We use MMDetection\footnote{\url{https://github.com/open-mmlab/mmdetection}} \cite{mmdetection} as the detector framework with 8 Nvidia GTX 1080Ti GPUs. Some other important settings of detectors are shown in Table \ref{table1}.

\noindent
\textbf{Evaluation Metric.}
For underwater image enhancement, since the corresponding ground truth images (clear images) are difficult to obtain, we use 8 commonly-used no-reference image quality assessment metrics including average gradient (AG) \cite{zhang2019single}, edge intensity (EI) \cite{azmi2019natural}, information entropy (IE) \cite{zhang2019single}, underwater image quality measure (UIQM) \cite{yang2015underwater}, and underwater color image quality evaluation metric (UCIQE) \cite{panetta2015human} to quantify the performance of different underwater image enhancement algorithms. 
A higher AG score of an image is indicative of a superior level of detail and enhanced image clarity.
A higher IE score indicates a greater amount of information contained within the image.
A higher EI score indicates the presence of more edges and texture information within the image, thereby indicating a higher quality of the image.
UCIQE evaluates underwater image quality by color density, saturation, and contrast. 
UIQM is a comprehensive underwater image evaluation index, which is the weighted sum of underwater image colorfulness measure (UICM), underwater image sharpness measure (UISM) and underwater image contrast measure (UIConM): $UIQM = {c_1} \times UICM + {c_2} \times UISM + {c_3} \times UIConM$. We set $c_1=0.0282$, $c_2=0.2953$, and $c_3=3.5753$ by default \cite{panetta2015human}. A higher UCIQE or UIQM score indicates the result has better colorfulness, sharpness, and contrast. The enhancement metrics are all calculated on the original image size without any resizing operation.
The performance of underwater object detection is measured by the commonly used COCO Average Precision (AP) \cite{lin2014microsoft}. 
In detail, AP is averaged over multiple IoU values ($0.5:0.05:0.95$).
Additionally, for the convenience of observation and analysis, we added a subscript indicating the ranking of each metric in the tables in our paper and provided an Average Ranking for the ranking statistics of all metrics.

\begin{table}[!t]    
  \centering
  \caption{\textbf{The training settings of each detector.} (Schd: Schedule. LR: Learning Rate)}
  \begin{tabular}{c|c|c|c|c}
  \hline
  \textbf{Detectors}      & \textbf{Schd} & \textbf{LR} & \textbf{\begin{tabular}[c]{@{}c@{}}Warmup\\ Iters\end{tabular}} & \textbf{\begin{tabular}[c]{@{}c@{}}Batch\\ Size\end{tabular}} \\ \hline
  Faster R-CNN \cite{ren2015faster}  & 1x        & 0.02          & 500       & 16   \\
  Cascade R-CNN \cite{cai2018cascade} & 1x        & 0.02          & 500       & 16  \\
  RetinaNet \cite{lin2017focal}    & 1x        & 0.01          & 1000      & 16   \\
  FCOS \cite{tian2019fcos}         & 1x        & 0.01          & 1000      & 16   \\
  ATSS \cite{zhang2020bridging}         & 1x        & 0.01          & 500       & 16   \\
  TOOD \cite{feng2021tood}          & 1x        & 0.01          & 500       & 16   \\
  SSD \cite{liu2016ssd}          & 120e      & 0.002         & 1000      & 64   \\ \hline
\end{tabular}
\label{table1}
\end{table}

\begin{figure*}[!t]
  \centering
  \includegraphics[width=1\linewidth]{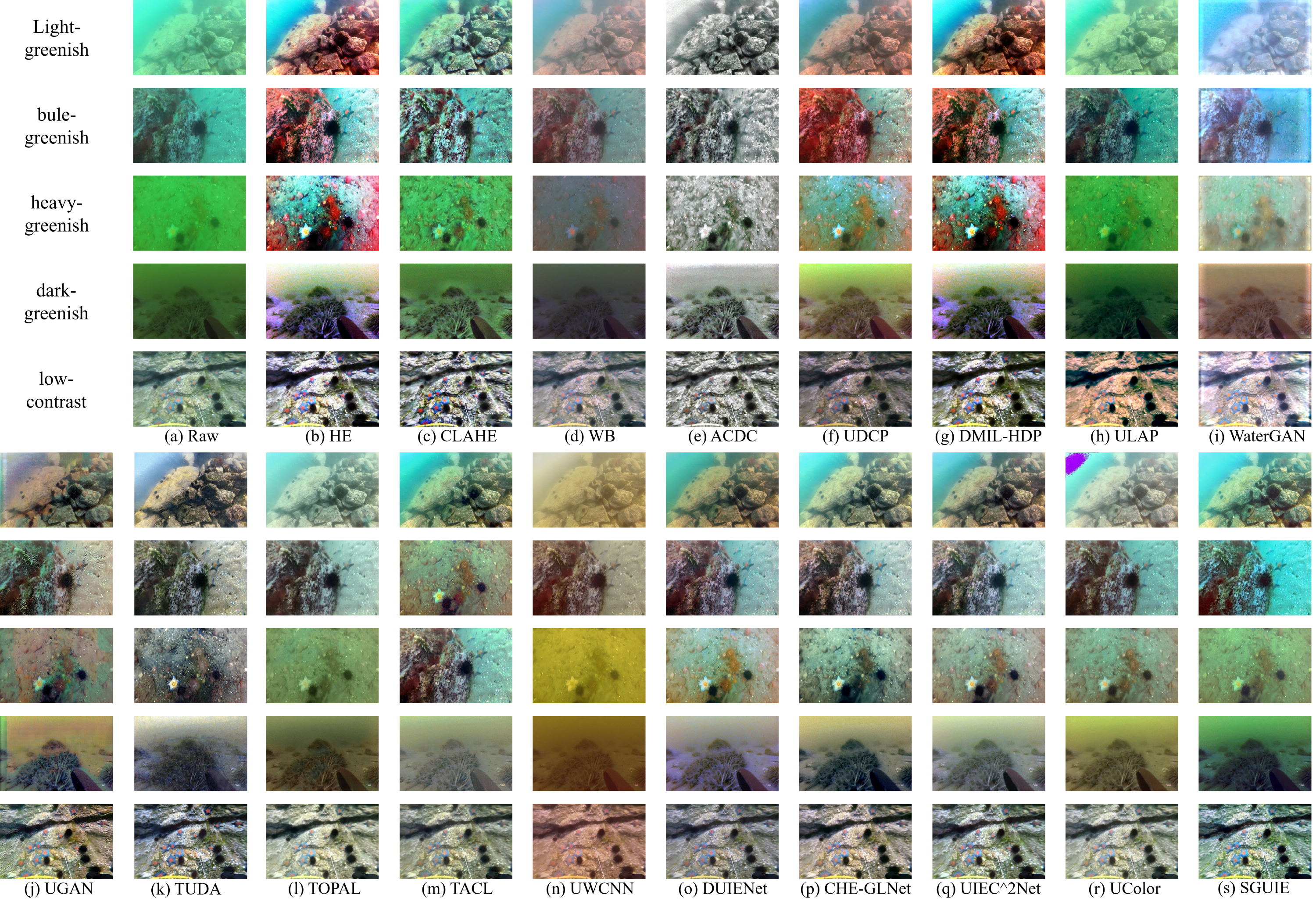}
  \caption{\textbf{Qualitative comparisons of different underwater image enhancement algorithms.} (a) Five raw underwater images sampled from underwater object detection dataset URPC2020 \cite{liu2021dataset}. From top to bottom are the samples of light-greenish, blue-greenish, heavy-greenish, dark-greenish, and hazy underwater images. The results enhanced by (b) HE \cite{hummel1977image}, (c) CLAHE \cite{zuiderveld1994contrast}, (d) WB \cite{ebner2007color}, (e) ACDC \cite{zhang2022underwater}, (f) UDCP \cite{drews2016underwater}, (g) DMIL-HDP \cite{li2016mlhp}, (h) ULAP \cite{song2018rapid}, (i) WaterGAN \cite{li2017watergan}, (j) UGAN \cite{fabbri2018enhancing}, (k) TUDA \cite{wang2023domain}, (l) TOPAL \cite{jiang2022target}, (m) TACL \cite{liu2022twin}, (n) UWCNN \cite{li2020underwater}, (o) DUIENet \cite{li2019underwater}, (p) CHE-GLNet \cite{fu2020underwater}, (q) UIEC\^{}2Net \cite{wang2021uiec}, (r) UColor \cite{li2021underwater}, and (s) SGUIE \cite{qi2022sguie}.}
  \label{fig2}
\end{figure*}

\begin{table*}[!t]    
\centering
\caption{\textbf{Quantitative scores of different underwater image enhancement algorithms in terms of the average values of AG \cite{zhang2019single}, EI \cite{azmi2019natural}, IE \cite{zhang2019single}, UIQM \cite{yang2015underwater}, UICM \cite{yang2015underwater}, UISM \cite{yang2015underwater}, UIConM \cite{yang2015underwater}, and UCIQE \cite{panetta2015human} on underwater object detection dataset.} The highest performance in \textcolor{rred}{red}, while the second best value and the third best value are in  \textcolor{bblue}{blue} and \textcolor{yyellow}{yellow}, respectively.}
	
\begin{tabular}{c|llllllll|c}
\hline
\textbf{Methods}                    & 
\multicolumn{1}{c}{\textbf{AG} $\uparrow$}  & \multicolumn{1}{c}{\textbf{EI} $\uparrow$}  &\multicolumn{1}{c}{\textbf{IE} $\uparrow$}  &\multicolumn{1}{c}{\textbf{UIQM} $\uparrow$} &\multicolumn{1}{c}{\textbf{UICM} $\uparrow$} & \multicolumn{1}{c}{\textbf{UISM} $\uparrow$} &\multicolumn{1}{c}{\textbf{UIConM} $\uparrow$} &\multicolumn{1}{c|}{\textbf{UCIQE} $\uparrow$}  & \textbf{\makecell{Average   \\  Ranking}}  \\ \hline
RAW                                              &{$1.205_{19}$}  &	{$12.720_{19}$}  &	{$6.876_{18}$}  &	{$1.644_{19}$}  &	{$0.330_{18}$}  &	{$1.240_{19}$}  &	{$0.354_{17}$}  &	{$0.406_{19}$}  &	18.500  \\
HE \cite{hummel1977image}                        &\cellcolor{rred}{$5.339_{1}$}  &	\cellcolor{rred}{$56.643_{1}$}  &	\cellcolor{rred}{$7.971_{1}$}  &	\cellcolor{rred}{$3.750_{1}$}  &	\cellcolor{rred}{$8.562_{1}$}  &	\cellcolor{yyellow}{$3.752_{3}$}  &	\cellcolor{yyellow}{$0.671_{3}$}  &	\cellcolor{bblue}{$0.675_{2}$}  &	\cellcolor{rred}{1.625}  \\
CLAHE \cite{zuiderveld1994contrast}              &{$3.561_{6}$}  &	{$37.732_{6}$}  &	{$7.301_{12}$}  &	{$3.136_{7}$}  &	{$1.931_{13}$}  &	{$3.049_{7}$}  &	{$0.610_{6}$}  &	{$0.487_{13}$}  &	8.750  \\
WB \cite{ebner2007color}                         &{$1.288_{18}$}  &	{$13.629_{18}$}  &	{$6.182_{19}$}  &	{$1.703_{18}$}  &	{$2.233_{10}$}  &	{$1.301_{18}$}  &	{$0.351_{18}$}  &	{$0.434_{17}$}  &	17.000  \\
ACDC \cite{zhang2022underwater}                  &{$4.514_{4}$}  &	{$47.866_{4}$}  &	\cellcolor{yyellow}{$7.629_{3}$}  &	{$3.400_{6}$}  &	{$0.870_{16}$}  &	{$3.566_{5}$}  &	{$0.649_{4}$}  &	{$0.517_{11}$}  &	6.625  \\
UDCP \cite{drews2016underwater}                  &{$2.692_{10}$}  &	{$28.619_{10}$}  &	{$7.328_{11}$}  &	{$2.651_{11}$}  &	\cellcolor{yyellow}{$4.136_{3}$}  &	{$2.512_{12}$}  &	{$0.501_{12}$}  &	{$0.553_{8}$}  &	9.625  \\
DMIL-HDP \cite{li2016mlhp}                       &\cellcolor{bblue}{$4.979_{2}$}  &	\cellcolor{bblue}{$52.878_{2}$}  &	\cellcolor{bblue}{$7.902_{2}$}  &	\cellcolor{bblue}{$3.706_{2}$}  &	\cellcolor{bblue}{$7.716_{2}$}  &	{$3.625_{4}$}  &	\cellcolor{bblue}{$0.676_{2}$}  &	\cellcolor{rred}{$0.680_{1}$}  &	\cellcolor{bblue}{2.125}  \\
ULAP \cite{song2018rapid}                        &{$1.982_{13}$}  &	{$20.874_{14}$}  &	{$7.074_{17}$}  &	{$2.361_{13}$}  &	{$1.589_{15}$}  &	{$1.909_{14}$}  &	{$0.490_{13}$}  &	{$0.476_{15}$}  &	14.250  \\
WaterGAN \cite{li2017watergan}                   &{$1.710_{15}$}  &	{$18.443_{16}$}  &	{$7.190_{15}$}  &	{$1.766_{17}$}  &	{$2.039_{11}$}  &	{$1.727_{15}$}  &	{$0.335_{19}$}  &	{$0.479_{14}$}  &	15.250  \\
UGAN \cite{fabbri2018enhancing}                  &\cellcolor{yyellow}{$4.739_{3}$}  &	\cellcolor{yyellow}{$48.649_{3}$}  &	{$7.257_{13}$}  &	{$3.470_{5}$}  &	{$2.457_{7}$}  &	\cellcolor{bblue}{$3.789_{2}$}  &	{$0.638_{5}$}  &	{$0.558_{6}$}  &	5.500  \\
TUDA \cite{wang2023domain}                       &{$4.028_{5}$}  &	{$43.227_{5}$}  &	{$7.552_{5}$}  &	{$3.530_{4}$}  &	{$1.655_{14}$}  &	{$3.511_{6}$}  &	\cellcolor{rred}{$0.684_{1}$}  &	{$0.560_{5}$}  &	5.625  \\
TOPAL \cite{jiang2022target}                     &{$1.676_{16}$}  &	{$18.616_{15}$}  &	{$7.159_{16}$}  &	{$1.998_{15}$}  &	{$0.786_{17}$}  &	{$1.714_{16}$}  &	{$0.411_{15}$}  &	{$0.473_{16}$}  &	15.750  \\
TACL \cite{liu2022twin}                          &{$3.327_{7}$}  &	{$35.742_{7}$}  &	{$7.458_{6}$}  &	\cellcolor{yyellow}{$3.605_{3}$}  &	{$3.0145_{6}$}  &	\cellcolor{rred}{$4.892_{1}$}  &	{$0.581_{7}$}  &	\cellcolor{yyellow}{$0.574_{3}$}  &	\cellcolor{yyellow}{5.000}  \\
UWCNN \cite{li2020underwater}                       &{$1.559_{17}$}  &	{$16.446_{17}$}  &	{$7.239_{14}$}  &	{$1.926_{16}$}  &	{$0.009_{19}$}  &	{$1.651_{17}$}  &	{$0.407_{16}$}  &	{$0.433_{18}$}  &	16.750  \\
DUIENet \cite{li2019underwater}                  &{$2.568_{12}$}  &	{$27.398_{12}$}  &	{$7.338_{9}$}  &	{$2.697_{10}$}  &	{$3.144_{4}$}  &	{$2.533_{11}$}  &	{$0.520_{10}$}  &	{$0.556_{7}$}  &	9.375  \\
CHE-GLNet \cite{fu2020underwater}                &{$3.246_{8}$}  &	{$33.936_{8}$}  &	{$7.618_{4}$}  &	{$2.920_{8}$}  &	{$2.265_{9}$}  &	{$2.843_{8}$}  &	{$0.564_{8}$}  &	{$0.569_{4}$}  &	7.125  \\
UIEC\textasciicircum{}2Net \cite{wang2021uiec}   &{$2.677_{11}$}  &	{$28.313_{11}$}  &	{$7.437_{7}$}  &	{$2.631_{12}$}  &	{$2.278_{8}$}  &	{$2.538_{10}$}  &	{$0.508_{11}$}  &	{$0.552_{9}$}  &	9.875  \\
UColor \cite{li2021underwater}                   &{$1.889_{14}$}  &	{$21.108_{13}$}  &	{$7.332_{10}$}  &	{$2.084_{14}$}  &	{$1.954_{12}$}  &	{$1.985_{13}$}  &	{$0.437_{14}$}  &	{$0.503_{12}$}  &	12.750  \\
SGUIE \cite{qi2022sguie}                         &{$2.739_{9}$}  &	{$29.825_{9}$}  &	{$7.396_{8}$}  &	{$2.740_{9}$}  &	{$3.066_{5}$}  &	{$2.564_{9}$}  &	{$0.531_{9}$}  &	{$0.532_{10}$}  &	8.500  \\
 \hline
	\end{tabular}
\label{table2}
\end{table*}

\begin{table*}[!t]    
	\centering
	\caption{\textbf{The AP result (\%) of Faster R-CNN \cite{ren2015faster}, Cascade R-CNN \cite{cai2018cascade}, RetinaNet \cite{lin2017focal}, FCOS \cite{tian2019fcos}, ATSS \cite{zhang2020bridging}, TOOD \cite{feng2021tood}, and SSD \cite{liu2016ssd} that training with raw and different enhanced URPC2020 dataset \cite{liu2021dataset} and testing with the same domain.} The highest performance in \textcolor{rred}{red}, the second best value is in \textcolor{bblue}{blue}, and the third best value is in \textcolor{yyellow}{yellow}. We add subscripts after the AP values, which represent the current ranking of the detector performance.}
	\resizebox{\linewidth}{!}{
	\begin{tabular}{c|ccccccc|c}
			\hline
\textbf{Methods}           & \textbf{\makecell{  Faster  \\   R-CNN\cite{ren2015faster}}} & \textbf{\makecell{  Cascade   \\  R-CNN  \cite{cai2018cascade}}} & \textbf{RetinaNet \cite{lin2017focal}} & \textbf{FCOS \cite{tian2019fcos}} & \textbf{ATSS \cite{zhang2020bridging}} & \textbf{TOOD \cite{feng2021tood}} & \textbf{SSD \cite{liu2016ssd}} & \textbf{\makecell{Average   \\  Ranking}} \\ \hline
RAW                                              & \cellcolor{rred}{$43.5_{1}$}	        &	\cellcolor{rred}{$44.3_{1}$}	&	\cellcolor{rred}{$40.7_{1}$}	&	\cellcolor{rred}{$41.4_{1}$}	&	\cellcolor{rred}{$44.8_{1}$}	&	\cellcolor{rred}{$45.4_{1}$}	&	\cellcolor{rred}{$35.2_{1}$}	&	\cellcolor{rred}{1.00} 	\\
HE \cite{hummel1977image}                        & {$41.1_{15}$}	&	{$41.9_{15}$}	&	{$38.1_{15}$}	&	{$38.1_{11}$}	&	{$42.2_{14}$}	&	{$42.5_{16}$}	&	{$32.6_{15}$}	&	14.43 	\\
CLAHE \cite{zuiderveld1994contrast}              & {$42.8_{6}$}	    &	{$43.7_{4}$}	&	{$40.2_{5}$}	&	{$39.5_{8}$}	&	\cellcolor{yyellow}{$44.1_{3}$}	&	\cellcolor{yyellow}{$44.7_{3}$}	&	\cellcolor{yyellow}{$35.0_{3}$}	&	4.57 	\\
WB \cite{ebner2007color}                         & {$42.6_{7}$}	    &	{$43.7_{4}$}	&	\cellcolor{yyellow}{$40.3_{3}$}	    &	{$30.5_{18}$}	&	\cellcolor{yyellow}{$44.1_{3}$}	&	{$44.4_{4}$}	&	{$34.3_{11}$}	&	7.14 	\\
ACDC \cite{zhang2022underwater}                  & {$41.4_{13}$}	&	{$42.3_{13}$}	&	{$38.5_{13}$}	&	{$37.0_{15}$}	&	{$42.7_{13}$}	&	{$43.2_{13}$}	&	{$33.9_{12}$}	&	13.14 	\\
UDCP \cite{drews2016underwater}                  & {$42.5_{9}$}	    &	{$43.2_{11}$}	&	{$39.0_{11}$}	&	{$40.4_{5}$}	&	{$43.6_{7}$}	&	{$44.3_{7}$}	&	{$34.4_{9}$}	&	8.43 	\\
DMIL-HDP \cite{li2016mlhp}                       & {$41.2_{14}$}	&	{$42.0_{14}$}   &	{$38.5_{13}$}	&	{$37.9_{12}$}	&	{$42.0_{16}$}	&	{$42.6_{15}$}	&	{$32.6_{15}$}	&	14.14 	\\
ULAP \cite{song2018rapid}                        & \cellcolor{bblue}{$43.1_{2}$}	    &	\cellcolor{bblue}{$44.1_{2}$}	&	{$40.0_{6}$}	&	{$37.5_{14}$}	&	{$44.0_{5}$}	&	\cellcolor{bblue}{$44.8_{2}$}	&	{$34.4_{9}$}	&	5.71 	\\
WaterGAN \cite{li2017watergan}                   & {$41.0_{16}$}	&	{$41.8_{16}$}	&	{$37.4_{17}$}	&	{$20.8_{19}$}	&	{$42.2_{14}$}	&	{$42.4_{17}$}	&	{$31.5_{17}$}	&	16.57 	\\
UGAN \cite{fabbri2018enhancing}                  & {$39.0_{19}$}	&	{$39.8_{19}$}	&	{$36.3_{19}$}	&	{$36.1_{16}$}	&	{$40.5_{19}$}	&	{$40.8_{19}$}	&	{$30.1_{19}$}	&	18.57 	\\
TUDA \cite{wang2023domain}                       & {$39.8_{18}$}	&	{$40.6_{18}$}	&	{$36.7_{18}$}	&	{$37.8_{13}$}	&	{$41.1_{18}$}	&	{$41.1_{18}$}	&	{$30.6_{18}$}	&	17.29 	\\
TOPAL \cite{jiang2022target}                     & {$43.0_{4}$}	    &	{$43.6_{6}$}	&	\cellcolor{bblue}{$40.4_{2}$}	&	\cellcolor{bblue}{$40.7_{2}$}	&	\cellcolor{bblue}{$44.2_{2}$}	&	{$44.4_{4}$}	&	{$34.9_{5}$}	&	\cellcolor{yyellow}{3.57} 	\\
TACL \cite{liu2022twin}                          & {$40.9_{17}$}	&	{$41.7_{17}$}	&	{$38.1_{15}$}	&	{$38.9_{10}$}	&	{$41.9_{17}$}	&	{$42.7_{14}$}	&	{$33.0_{14}$}	&	14.86 	\\
UWCNN \cite{li2020underwater}                       & {$41.9_{12}$}	&	{$42.7_{12}$}	&	{$38.9_{12}$}	&	{$31.2_{17}$}	&	{$42.9_{12}$}	&	{$44.0_{11}$}	&	{$33.5_{13}$}	&	12.71 	\\
DUIENet \cite{li2019underwater}                  & {$42.3_{11}$}	&	{$43.4_{7}$}	&	{$39.8_{7}$}	&	{$39.5_{8}$}	&	{$43.2_{11}$}	&	{$43.9_{12}$}	&	{$34.8_{6}$}	&	8.86 	\\
CHE-GLNet \cite{fu2020underwater}                & {$42.6_{7}$}	    &	{$43.4_{7}$}	&	{$39.6_{8}$}	&	{$40.0_{7}$}	&	{$43.4_{10}$}	&	{$44.1_{10}$}	&	{$34.7_{8}$}	&	8.14 	\\
UIEC\textasciicircum{}2Net \cite{wang2021uiec}   & {$42.9_{5}$}	    &	{$43.4_{7}$}	&	{$39.6_{8}$}	&	{$40.1_{6}$}	&	{$43.6_{7}$}	&	{$44.3_{7}$}	&	{$34.8_{6}$}	&	6.57 	\\
UColor \cite{li2021underwater}                   & {$42.4_{10}$}	&	{$43.4_{7}$}	&	{$39.5_{10}$}	&	{$40.5_{4}$}	&	{$43.6_{7}$}	&	{$44.3_{7}$}	&	\cellcolor{bblue}{$35.1_{2}$}	&	6.71 	\\
SGUIE \cite{qi2022sguie}                         & \cellcolor{bblue}{$43.1_{2}$}	    &	\cellcolor{yyellow}{$44.0_{3}$}	&	\cellcolor{yyellow}{$40.3_{3}$}	&	\cellcolor{bblue}{$40.7_{2}$}	&	{$43.9_{6}$}	&	{$44.4_{4}$}	&	\cellcolor{yyellow}{$35.0_{3}$}	&	\cellcolor{bblue}{3.29} 	\\
\hline
	\end{tabular}}
	\label{table3}
\end{table*}

\subsection{Analysis of the Enhanced Results}

\noindent
\textbf{Qualitative Evaluation.} We simply divide the underwater images of URPC2020 \cite{liu2021dataset} into five categories according to image appearance: light-greenish images, blue-greenish images, heavy-greenish images, dark-greenish images, and low-contrast images. The results enhanced by different underwater image enhancement algorithms are shown in Fig. \ref{fig2}.

Due to the nature of light propagation, in most cases, the red light first disappears in water, followed by the green light and then the blue light.
Such selective attenuation results in greenish and bluish underwater images, such as the raw underwater images in Fig. \ref{fig2}(a). Color deviation seriously affects the visual quality of underwater images. 
For the images of the first three categories, HE \cite{hummel1977image}, UDCP \cite{zhang2022underwater}, and DMIL-HDP \cite{li2016mlhp} introduce reddish or purplish color cast due to the inaccurate color correction methods. 
CLAHE \cite{zuiderveld1994contrast}, ULAP \cite{li2016mlhp}, and SGUIE \cite{qi2022sguie} have a less positive effect on the enhancement of greenish underwater images, while UWCNN \cite{li2020underwater} exacerbates the color distortion of yellowish images. 
WB \cite{ebner2007color} and ACDC \cite{zhang2022underwater} can successfully remove the color cast but WB leads to the low contrast problem and ACDC brings a low saturation problem. 
DUIENet \cite{li2019underwater}, CHE-GLNet \cite{fu2020underwater}, UIEC\textasciicircum{}2Net \cite{wang2021uiec}, TUDA \cite{wang2023domain}, TOPAL \cite{jiang2022target}, TACL \cite{liu2022twin}, and UColor \cite{li2021underwater} remove the color cast well and have the best visual quality. 
WaterGAN \cite{li2017watergan} and UGAN \cite{fabbri2018enhancing} suffer from the color blocking problem and have poor edge detail. 
For the dark-greenish underwater images, the situation is similar to the above, except that the DUIENet \cite{li2019underwater}, CHE-GLNet \cite{fu2020underwater}, and UIEC\textasciicircum{}2Net \cite{wang2021uiec} introduce some color casts and TUDA \cite{wang2023domain} introduces edge blur problem.
For the hazy underwater images as shown in the last row of Fig. \ref{fig2}(a). HE \cite{hummel1977image}, CLAHE \cite{zuiderveld1994contrast}, ACDC \cite{zhang2022underwater}, UDCP \cite{zhang2022underwater}, DMIL-HDP \cite{li2016mlhp}, WaterGAN \cite{li2017watergan}, UGAN \cite{fabbri2018enhancing}, TUDA \cite{wang2023domain}, TOPAL \cite{jiang2022target}, TACL \cite{liu2022twin}, DUIENet \cite{li2019underwater}, CHE-GLNet \cite{fu2020underwater}, UIEC\^{}2Net \cite{wang2021uiec}, UColor \cite{li2021underwater}, and SGUIE \cite{qi2022sguie} significantly remove the effects of haze on the underwater images, while WB \cite{ebner2007color} suffers from low contrast problem. ULAP \cite{li2016mlhp} and UWCNN \cite{li2020underwater} introduce reddish color deviations.

\noindent
\textbf{Quantitative Evaluation.} 
Table \ref{table2} reports the averaged quantitative scores of different underwater image enhancement algorithms on the underwater object detection dataset. 
HE \cite{hummel1977image} obtains the highest AG \cite{zhang2019single}, EI \cite{azmi2019natural}, IE \cite{zhang2019single}, UIQM \cite{yang2015underwater}, and UICM values, the second best UCIQE \cite{panetta2015human} value, and the third best UISM and UIConM value, which achieves the best rank in objective evaluation metrics.
Meanwhile, DMIL-HDP \cite{li2016mlhp} obtains the highest UCIQE and the second best AG, EI, IE, UIQM, UICM, and UIConM values, which achieves the second best rank in quantitative scores.
TUDA \cite{wang2023domain} and TACL \cite{liu2022twin} obtain the highest UIConM and UISM value, respectively.
In comparison, the raw domain has the lowest  scores, followed by WB \cite{ebner2007color}, UWCNN \cite{li2020underwater}, TOPAL \cite{jiang2022target}, and WaterGAN \cite{li2017watergan}. 
Compared with qualitative analysis, we obtained a similar conclusion to \cite{liu2020real} that \textit{the quantitative scores are not always consistent with the image quality perceived by human vision, which implies there is a gap between the current image quality assessment metrics and the visual quality for underwater images.}

\begin{table*}[!t]    
	\centering
	\caption{\textbf{The AP result (\%) of Faster R-CNN \cite{ren2015faster} with different backbones that training with 1x \& 2x schedules on raw and different different enhanced URPC2020 dataset \cite{liu2021dataset} and testing with the same domain.} The highest performance in \textcolor{rred}{red}, the second best value is in \textcolor{bblue}{blue}, and the third best value is in \textcolor{yyellow}{yellow}.}
	\resizebox{\linewidth}{!}{
	\begin{tabular}{c|ccccc|ccccc}
\hline
\textit{\textbf{}}                             & \multicolumn{5}{c|}{\textit{\textbf{1$\times$ }}}                                                                                                                                                                                                                                                                      & \multicolumn{5}{c}{\textit{\textbf{2$\times$}}}                                                                                                                                                                                                                                                                        \\ \hline
\textbf{Methods}                               & \textbf{ResNet50}           & \textbf{ResNet101}          & \textbf{\begin{tabular}[c]{@{}c@{}}ResNext101\\ (32$\times$4d)\end{tabular}} & \multicolumn{1}{c|}{\textbf{\begin{tabular}[c]{@{}c@{}}ResNext101\\ (64$\times$4d)\end{tabular}}} & \textbf{\begin{tabular}[c]{@{}c@{}}Average\\ Rank\end{tabular}} & \textbf{ResNet50}            & \textbf{ResNet101}          & \textbf{\begin{tabular}[c]{@{}c@{}}ResNext101\\ (32$\times$4d)\end{tabular}} & \multicolumn{1}{c|}{\textbf{\begin{tabular}[c]{@{}c@{}}ResNext101\\ (64$\times$4d)\end{tabular}}} & \textbf{\begin{tabular}[c]{@{}c@{}}Average\\ Ranking\end{tabular}} \\ \hline
RAW                                              & \cellcolor{rred}{$43.5_{1}$}	&	\cellcolor{rred}{$44.8_{1}$}	&	\cellcolor{rred}{$44.6_{1}$}	&	\cellcolor{rred}{$45.3_{1}$}	&	\multicolumn{1}{|c|}{\cellcolor{rred}{1.00}}   & 	\cellcolor{rred}{$43.1_{1}$}	&	\cellcolor{rred}{$43.2_{1}$}	&	\cellcolor{rred}{$42.4_{1}$}	&	\cellcolor{rred}{$42.3_{1}$}	&	\multicolumn{1}{|c}{\cellcolor{rred}{1.00}}	\\
HE \cite{hummel1977image}                        & {$41.1_{15}$}	&	{$42.7_{13}$}	&	{$42.4_{15}$}	&	{$42.7_{14}$}	                                                            &	\multicolumn{1}{|c|}{14.25}  &  {$40.3_{14}$}	&	{$40.4_{16}$}	&	{$40.0_{13}$}	&	{$40.1_{15}$}	                                                                    &	\multicolumn{1}{|c}{14.50} 	\\
CLAHE \cite{zuiderveld1994contrast}              & {$42.8_{6}$}	    &	{$44.1_{4}$}	&	{$43.4_{12}$}	&	\cellcolor{yyellow}{$44.4_{3}$}	                                            &	\multicolumn{1}{|c|}{6.25}   & 	{$41.5_{8}$}	&	{$41.8_{6}$}	&	{$41.4_{5}$}	&	\cellcolor{bblue}{$41.6_{2}$}	                                                &	\multicolumn{1}{|c}{5.25}	\\
WB \cite{ebner2007color}                         & {$42.6_{7}$}	    &	{$43.9_{7}$}	&	\cellcolor{yyellow}{$44.2_{3}$}	&	\cellcolor{yyellow}{$44.4_{3}$}	                            &	\multicolumn{1}{|c|}{5.00}   & 	{$41.5_{8}$}	&	\cellcolor{yyellow}{$42.5_{3}$}	&	\cellcolor{yyellow}{$41.6_{3}$}	&	{$41.4_{5}$}	                                &	\multicolumn{1}{|c}{4.75}	\\
ACDC \cite{zhang2022underwater}                  & {$41.4_{13}$}	&	{$42.4_{16}$}	&	{$42.7_{13}$}	&	{$43.4_{13}$}	                                                            &	\multicolumn{1}{|c|}{13.75}  &  {$40.2_{15}$}	&	{$40.3_{17}$}	&	{$39.8_{14}$}	&	{$40.3_{13}$}	                                                                &	\multicolumn{1}{|c}{14.75} 	\\
UDCP \cite{drews2016underwater}                  & {$42.5_{9}$}	    &	{$43.9_{7}$}	&	{$43.8_{9}$}	&	{$44.3_{6}$}	                                                            &	\multicolumn{1}{|c|}{7.75}   & 	{$41.6_{5}$}	&	{$41.6_{9}$}	&	{$41.1_{9}$}	&	{$41.2_{6}$}	                                                                &	\multicolumn{1}{|c}{7.25}	\\
DMIL-HDP \cite{li2016mlhp}                       & {$41.2_{14}$}	&	{$42.5_{15}$}	&	{$42.5_{14}$}	&	{$42.6_{15}$}	                                                            &	\multicolumn{1}{|c|}{14.50}  &  {$39.9_{17}$}	&	{$40.6_{14}$}	&	{$39.8_{14}$}	&	{$40.3_{13}$}	                                                                &	\multicolumn{1}{|c}{14.50} 	\\
ULAP \cite{song2018rapid}                        & \cellcolor{bblue}{$43.1_{2}$}	&	\cellcolor{bblue}{$44.5_{2}$}	&	{$44.0_{6}$}	&	\cellcolor{bblue}{$44.5_{2}$}	            &	\multicolumn{1}{|c|}{\cellcolor{yyellow}{3.00}}   & 	\cellcolor{bblue}{$42.2_{2}$}	&	\cellcolor{bblue}{$42.6_{2}$}	&	\cellcolor{bblue}{$41.8_{2}$}	&	\cellcolor{yyellow}{$41.5_{3}$}	&	\multicolumn{1}{|c}{\cellcolor{bblue}{2.25}}	\\
WaterGAN \cite{li2017watergan}                   & {$41.0_{16}$}    &	{$41.9_{17}$}	&	{$42.3_{17}$}	&	{$42.3_{17}$}	                                                            &	\multicolumn{1}{|c|}{16.75}  &  {$40.5_{13}$}	&	{$40.5_{15}$}	&	{$39.8_{14}$}	&	{$39.5_{17}$}	                                                                &	\multicolumn{1}{|c}{14.75} 	\\
UGAN \cite{fabbri2018enhancing}                  & {$39.0_{19}$}    &	{$40.3_{19}$}	&	{$40.8_{19}$}	&	{$41.2_{19}$}	                                                            &	\multicolumn{1}{|c|}{19.00}  &  {$38.4_{19}$}	&	{$38.9_{18}$}	&	{$38.0_{19}$}	&	{$38.5_{19}$}	                                                                    &	\multicolumn{1}{|c}{18.75} 	\\
TUDA \cite{wang2023domain}                       & {$39.8_{18}$}	&	{$40.8_{18}$}	&	{$41.0_{18}$}	&	{$41.4_{18}$}	                                                                &	\multicolumn{1}{|c|}{18.00}  &  {$38.5_{18}$}	&	{$38.8_{19}$}	&	{$38.1_{18}$}	&	{$38.8_{18}$}	                                                                &	\multicolumn{1}{|c}{18.25} 	\\
TOPAL \cite{jiang2022target}                     & {$43.0_{4}$}	    &	{$44.1_{4}$}	&	{$44.1_{4}$}	&	{$44.1_{8}$}	                                                            &	\multicolumn{1}{|c|}{5.00}   & 	{$42.0_{4}$}	&	{$42.3_{4}$}	&	{$41.4_{5}$}	&	{$41.2_{6}$}	                                                                    &	\multicolumn{1}{|c}{4.75}	\\
TACL \cite{liu2022twin}                          & {$40.9_{17}$}	&	{$42.6_{14}$}	&	{$42.4_{15}$}	&	{$42.5_{16}$}	                                                            &	\multicolumn{1}{|c|}{15.50}  &  {$40.2_{15}$}	&	{$40.7_{13}$}	&	{$39.6_{17}$}	&	{$39.8_{16}$}	                                                                &	\multicolumn{1}{|c}{15.25} 	\\
UWCNN \cite{li2020underwater}                       & {$41.9_{12}$}	&	{$43.4_{10}$}	&	{$43.5_{11}$}	&	{$43.9_{11}$}	                                                            &	\multicolumn{1}{|c|}{11.00}  &  {$41.6_{5}$}	&	{$41.8_{6}$}	&	{$41.3_{8}$}	&	{$40.9_{11}$}	                                                                &	\multicolumn{1}{|c}{7.50}	\\
DUIENet \cite{li2019underwater}                  & {$42.3_{11}$}	&	{$43.4_{10}$}	&	{$43.7_{10}$}	&	{$44.2_{7}$}	                                                            &	\multicolumn{1}{|c|}{9.50}   & 	{$41.4_{10}$}	&	{$41.4_{10}$}	&	{$41.0_{10}$}	&	{$41.0_{9}$}	                                                                        &	\multicolumn{1}{|c}{9.75}	\\
CHE-GLNet \cite{fu2020underwater}                & {$42.6_{7}$}	    &	{$43.5_{9}$}	&	{$43.9_{8}$}	&	{$43.8_{12}$}	                                                            &	\multicolumn{1}{|c|}{9.00}   & 	{$41.4_{10}$}	&	{$41.4_{10}$}	&	{$40.9_{11}$}	&	{$41.0_{9}$}	                                                                    &	\multicolumn{1}{|c}{10.00} 	\\
UIEC\textasciicircum{}2Net \cite{wang2021uiec}   & {$42.9_{5}$}	    &	{$43.4_{10}$}	&	{$44.0_{6}$}	&	{$44.1_{8}$}	                                                            &	\multicolumn{1}{|c|}{7.25}   & 	{$41.3_{12}$}	&	{$41.4_{10}$}	&	{$40.8_{12}$}	&	{$40.9_{11}$}	                                                                &	\multicolumn{1}{|c}{11.25} 	\\
UColor \cite{li2021underwater}                   & {$42.4_{10}$}	&	\cellcolor{yyellow}{$44.2_{3}$}	&	{$44.1_{4}$}	&	{$44.1_{8}$}	                                            &	\multicolumn{1}{|c|}{6.25}   & 	{$41.6_{5}$}	&	{$41.9_{5}$}	&	{$41.4_{5}$}	&	{$41.1_{8}$}	                                                                &	\multicolumn{1}{|c}{5.75}	\\
SGUIE \cite{qi2022sguie}                         & \cellcolor{bblue}{$43.1_{2}$}	&	{$44.1_{4}$}	&	\cellcolor{bblue}{$44.4_{2}$}	&	\cellcolor{yyellow}{$44.4_{3}$}	            &	\multicolumn{1}{|c|}{\cellcolor{bblue}{2.75}}   & 	\cellcolor{yyellow}{$42.1_{3}$}	&	{$41.8_{6}$}	&	\cellcolor{yyellow}{$41.6_{3}$}	&	\cellcolor{yyellow}{$41.5_{3}$}	                &	\multicolumn{1}{|c}{\cellcolor{yyellow}{3.75}}	\\
\hline
	\end{tabular}}
	\label{table4}
\end{table*}

\subsection{Analysis of the Underwater Object Detection Results}\label{exp_det}

\noindent
\textbf{Quantitative Analysis.} 
After gaining the underwater object detection images enhanced by different enhancement algorithms, we retrain the detectors separately and use the same corresponding images (the same domain) for testing. 
The results of different detectors with different domains are shown in Table \ref{table3}. We made several observations as follows:
1) surprisingly, all detectors retrained on the raw domain can achieve the highest AP values than the detectors retrained on other domains.
2) For the two-stage detectors (Faster R-CNN \cite{ren2015faster} and Cascade R-CNN \cite{cai2018cascade}), the detectors retrained on the ULAP \cite{song2018rapid} domain achieve the second best AP values when they are compared with the raw domain two-stage detectors with a decrease of 0.4\% and 0.2\%, respectively.
3) The detectors retrained on the SGUIE \cite{qi2022sguie}, TOPAL \cite{jiang2022target}, CLAHE \cite{zuiderveld1994contrast}, WB \cite{ebner2007color}, and UIEC\textasciicircum{}2Net \cite{wang2021uiec} domains also obtain satisfactory performance.
4) In contrast, the performance of the detectors retrained on the HE \cite{hummel1977image}, WaterGAN \cite{li2017watergan}, UGAN \cite{fabbri2018enhancing}, TUDA \cite{wang2023domain}, and TACL \cite{liu2022twin} domains degrade severely, only with the AP values of 41.1\%, 41.0\%, 39.0\%, 39.8\%, and 40.9\% on Faster R-CNN \cite{ren2015faster} and 41.9\%, 41.8\%, 39.8\%, 40.6\%, and 41.7\% on Cascade R-CNN, compared with the highest AP values 43.5\% and 44.3\%, respectively.
5) As for the one-stage detectors, the overall phenomenon is similar to the two-stage detectors. 
The RAW domain detectors can also achieve the highest AP values. The detectors retrained on the CLAHE \cite{zuiderveld1994contrast}, WB \cite{ebner2007color}, TOPAL \cite{jiang2022target}, and SGUIE \cite{qi2022sguie} can also achieve higher AP values. Moreover, the performance of the detectors retrained on the GAN-based methods, such as UGAN \cite{fabbri2018enhancing}, TUDA \cite{wang2023domain}, and TACL \cite{liu2022twin} degrade severely.
6) The performance of the CNN-based enhancement domain detectors is just behind the CLAHE \cite{zuiderveld1994contrast}, ULAP \cite{li2016mlhp}, and TOPAL \cite{jiang2022target} domain detectors, with an average ranking in the upper-middle (rankings from 3 to 8), while except TOPAL \cite{jiang2022target} the GAN-based enhancement domain detectors are not ideal, ranking at the bottom.

\begin{figure}[!th]     
  \centering
  \includegraphics[width=1\linewidth]{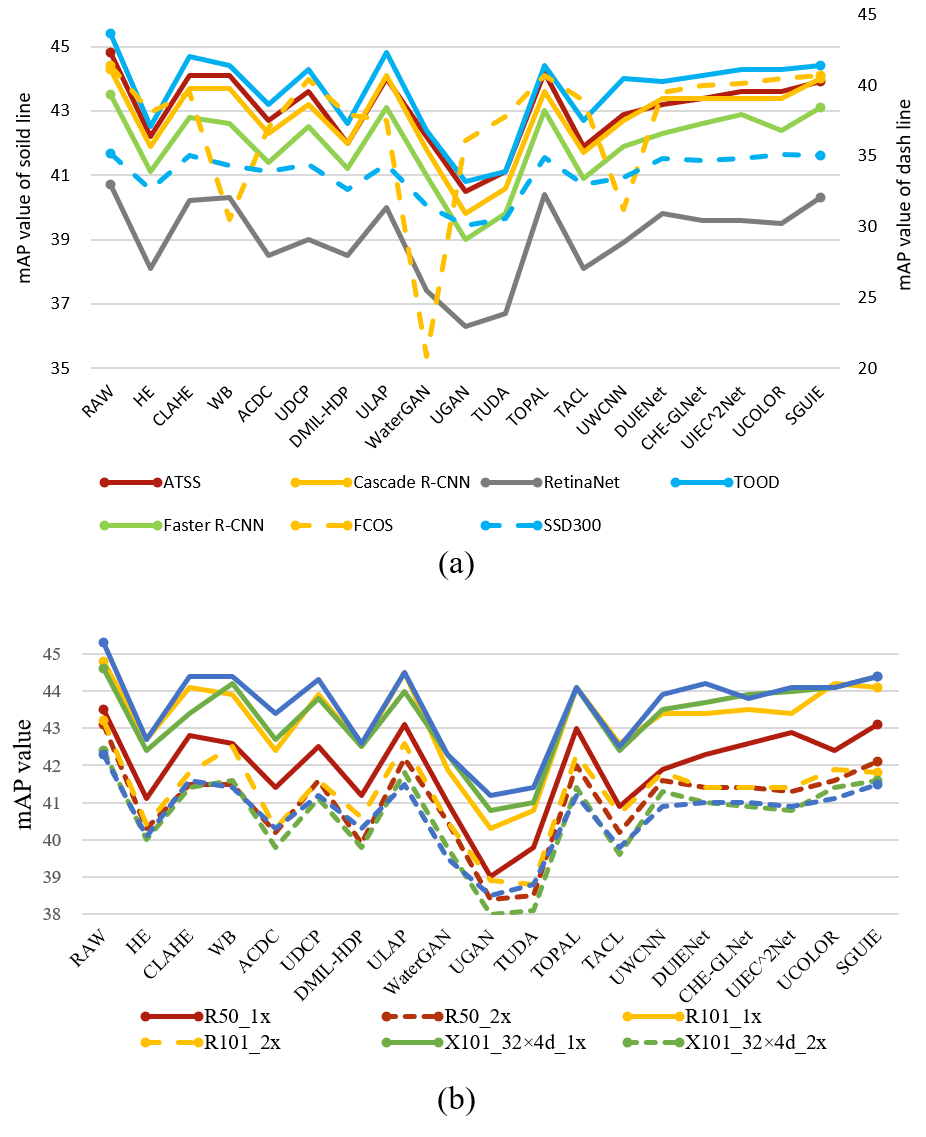}
  \caption{\textbf{The line graph of AP values.} (a) The line graph of AP values of Faster R-CNN \cite{ren2015faster}, Cascade R-CNN \cite{cai2018cascade}, RetinaNet \cite{lin2017focal}, FCOS \cite{tian2019fcos}, ATSS \cite{zhang2020bridging}, TOOD \cite{feng2021tood}, and SSD \cite{liu2016ssd}. (b) The line graph of AP values of Faster R-CNN \cite{ren2015faster} with different backbones and training schedules.}
  \label{fig3}
\end{figure}

In order to verify whether the backbone with stronger feature extraction ability or a longer training schedule can reduce the negative impact of underwater image enhancement on object detection.
We further perform two complementary experiments on Faster R-CNN \cite{ren2015faster}: one is to replace ResNet-50 \cite{he2016deep} with  stronger backbones (ResNet-101 and ResNext-101 \cite{xie2017aggregated}); another is to increase the training schedule to 2$\times$ (24 epochs). 
The results are shown in Table \ref{table4}.
We also find that the raw domain detectors achieve the best performance, which further validating that \textit{underwater image enhancement suppresses the performance of object detection}. 
SGUIE \cite{qi2022sguie} and ULAP \cite{song2018rapid} domain detectors achieve the second and third best performance. Compared with the raw domain detectors, there is a decrease of around 0.4. With the enhancement of the detector backbone, the gap between the enhanced domain and raw domain detector gradually widens.
Followed by the WB \cite{ebner2007color}, CLAHE \cite{zuiderveld1994contrast}, UIEC\textasciicircum{}2Net \cite{wang2021uiec}, and UColor \cite{li2021underwater} domain detectors, they also achieve a better performance.
In contrast, the performance of UGAN \cite{fabbri2018enhancing}, TUDA \cite{wang2023domain}, WaterGAN \cite{li2017watergan}, TACL \cite{liu2022twin}, HE \cite{hummel1977image}, and DMIL-HDP \cite{li2016mlhp} domain detectors degrade severely. 
Compared with the enhancement domain detectors, with the improvement of the backbone, the AP of raw domain detectors increases more significantly.
Moreover, the 2$\times$ schedule suffers the over-fitting problem, in which the AP value is lower than the corresponding 1$\times$  schedule detectors.

To better show the differences, we present the results of Table \ref{table3} and \ref{table4} as line graphs, which are shown in Fig. \ref{fig3}(a) and (b), respectively.
As shown in Fig. \ref{fig3}(a), on the raw dataset, TOOD \cite{feng2021tood} achieves the best results, followed by ATSS \cite{zhang2020bridging} and Cascade R-CNN \cite{cai2018cascade}.
Except for FCOS \cite{tian2019fcos}, other detectors follow a similar trend, i.e., the AP values of the detectors are correlated with the corresponding enhancement algorithms.
In summary, it can be visualized that the detector performance of each raw domain is higher than that of other enhanced domains.
A similar conclusion also exists in Fig. \ref{fig3}(b), in which the raw domain Faster R-CNN \cite{ren2015faster} with the backbone ResNeXt-101(64$\times$4d) \cite{xie2017aggregated} achieves the best performance followed by the backbones with ResNext-101(32$\times$4d), ResNet-101 \cite{he2016deep}, and ResNet-50 on 1$\times$ training schedule.
The line graph further shows that the detectors of different backbones and training schedules have the same trends in all domains.
That is, each detector retrained on the raw domain achieves the best performance. Moreover, each detector retrained on the ULAP \cite{song2018rapid}, TOPAL \cite{jiang2022target}, SGUIE \cite{qi2022sguie}, and UDCP \cite{drews2016underwater} domains also performs well. In contrast, the detectors retrained on the ACDC \cite{zhang2022underwater}, HE \cite{hummel1977image}, and DMIL-HDP \cite{li2016mlhp} domains obtain the low AP values. In addition, the detectors retrained on the UGAN \cite{fabbri2018enhancing} domain perform worst. 
Therefore, we obtain a preliminary conclusion that \textit{1) underwater image enhancement suppresses the performance of object detection and 2) the improvement of the network feature extraction capability cannot reduce the negative impact of the underwater image enhancement on object detection.}

\begin{table*}[htbp]    
  \centering
  \caption{The classification error, localization error, both Cls and Loc error, duplicate detection error, background error, missed GT error, FP error, FN error, and AP of Faster R-CNN \cite{ren2015faster} with different backbones and training schedules, RetinaNet \cite{lin2017focal}, Cascade R-CNN \cite{cai2018cascade}, FCOS \cite{tian2019fcos}, ATSS \cite{zhang2020bridging}, TOOD \cite{feng2021tood}, and SSD \cite{liu2016ssd}. The highest performance in \textcolor{rred}{red}, the second best value is in \textcolor{bblue}{blue}, and the third best value is in \textcolor{yyellow}{yellow}.}
  \resizebox{0.9\linewidth}{!}{
\begin{tabular}{c|cccccccc|c||cccccccc|c}
\hline
\hline
  & \multicolumn{9}{c||}{\textbf{Faster R-CNN (ResNet-50, 1x) \cite{ren2015faster}}}     & \multicolumn{9}{c}{\textbf{Faster R-CNN (ResNet-101, 1x) \cite{ren2015faster}}}           \\  \hline
\textbf{Methods}                               & \textbf{$E_{Cls} \downarrow$}    & \textbf{$E_{Loc} \downarrow$ }   & \textbf{$E_{Both} \downarrow$ }  & \textbf{$E_{Dupe} \downarrow$}   & \textbf{$E_{Bkg} \downarrow$}    & \textbf{$E_{Miss} \downarrow$}   & \textbf{$E_{FP} \downarrow$ }      & \textbf{$E_{FN} \downarrow$ }     & \textbf{AP (\%) $\uparrow$}          & \textbf{$E_{Cls} \downarrow$}     & \textbf{$E_{Loc} \downarrow$ }   & \textbf{$E_{Both} \downarrow$ }   & \textbf{$E_{Dupe} \downarrow$}    & \textbf{$E_{Bkg} \downarrow$}     & \textbf{$E_{Miss} \downarrow$}   & \textbf{$E_{FP} \downarrow$ }     & \textbf{$E_{FN} \downarrow$ }     & \textbf{AP (\%) $\uparrow$}          \\ \hline
RAW                                              &  {$1.40_{17}$}  &	{$3.78_{8}$}  &	{$0.22_{6}$}  &	\cellcolor{bblue}{$0.09_{2}$}  &	\cellcolor{bblue}{$5.27_{2}$}  &	\cellcolor{bblue}{$5.67_{2}$}  &	\cellcolor{rred}{$8.10_{1}$}  &	\cellcolor{rred}{$9.93_{1}$}  &	\cellcolor{rred}{$43.5_{1}$}  &  {$1.15_{7}$}  &	\cellcolor{bblue}{$3.50_{2}$}  &	{$0.20_{10}$}  &	\cellcolor{yyellow}{$0.06_{3}$}  &	\cellcolor{rred}{$4.70_{1}$}  &	\cellcolor{bblue}{$6.20_{2}$}  &	\cellcolor{rred}{$7.10_{1}$}  &	\cellcolor{rred}{$10.30_{1}$}  &	\cellcolor{rred}{$44.8_{1}$}  \\
HE \cite{hummel1977image}                        &  {$1.49_{19}$}  &	{$4.06_{15}$}  &	{$0.25_{17}$}  &	{$0.10_{11}$}  &	{$6.00_{16}$}  &	{$6.55_{13}$}  &	{$9.10_{16}$}  &	{$11.28_{14}$}  &	{$41.1_{15}$}  &  {$1.24_{12}$}  &	{$3.56_{4}$}  &	{$0.20_{10}$}  &	{$0.07_{9}$}  &	{$5.31_{16}$}  &	{$7.31_{15}$}  &	{$7.74_{13}$}  &	{$11.54_{14}$}  &	{$42.7_{13}$}  \\
CLAHE \cite{zuiderveld1994contrast}              &  {$1.27_{11}$}  &	{$3.68_{4}$}  &	\cellcolor{bblue}{$0.20_{2}$}  &	{$0.12_{18}$}  &	{$5.49_{6}$}  &	\cellcolor{yyellow}{$5.85_{3}$}  &	{$8.44_{7}$}  &	{$10.34_{5}$}  &	{$42.8_{6}$}  &  \cellcolor{bblue}{$1.01_{2}$}  &	{$3.76_{10}$}  &	\cellcolor{bblue}{$0.18_{2}$}  &	{$0.08_{16}$}  &	{$5.07_{11}$}  &	\cellcolor{rred}{$6.07_{1}$}  &	{$7.74_{13}$}  &	\cellcolor{bblue}{$10.37_{2}$}  &	{$44.1_{4}$}  \\
WB \cite{ebner2007color}                         &  {$1.23_{5}$}  &	{$4.04_{14}$}  &	\cellcolor{bblue}{$0.20_{2}$}  &	\cellcolor{bblue}{$0.09_{2}$}  &	\cellcolor{yyellow}{$5.40_{3}$}  &	{$5.91_{4}$}  &	{$8.43_{5}$}  &	\cellcolor{yyellow}{$10.26_{3}$}  &	{$42.6_{7}$}  &  {$1.50_{19}$}  &	{$3.80_{12}$}  &	{$0.21_{12}$}  &	{$0.09_{19}$}  &	{$4.86_{4}$}  &	{$6.73_{7}$}  &	{$7.35_{7}$}  &	{$10.89_{7}$}  &	{$43.9_{7}$}  \\
ACDC \cite{zhang2022underwater}                  &  {$1.26_{10}$}  &	{$4.35_{19}$}  &	{$0.23_{11}$}  &	{$0.10_{11}$}  &	{$5.86_{15}$}  &	{$6.57_{14}$}  &	{$9.14_{18}$}  &	{$11.48_{15}$}  &	{$41.4_{13}$}  &  {$1.22_{9}$}  &	{$4.14_{16}$}  &	{$0.23_{17}$}  &	{$0.07_{9}$}  &	{$5.27_{15}$}  &	{$7.22_{14}$}  &	{$8.04_{16}$}  &	{$12.02_{15}$}  &	{$42.4_{16}$}  \\
UDCP \cite{drews2016underwater}                  &  \cellcolor{rred}{$1.07_{1}$}  &	{$3.95_{10}$}  &	{$0.24_{14}$}  &	{$0.10_{11}$}  &	{$5.43_{5}$}  &	{$6.42_{9}$}  &	\cellcolor{yyellow}{$8.34_{3}$}  &	{$10.82_{10}$}  &	{$42.5_{9}$}  &  \cellcolor{rred}{$1.00_{1}$}  &	\cellcolor{bblue}{$3.50_{2}$}  &	{$0.19_{6}$}  &	\cellcolor{yyellow}{$0.06_{3}$}  &	\cellcolor{bblue}{$4.72_{2}$}  &	{$7.04_{13}$}  &	\cellcolor{bblue}{$7.11_{2}$}  &	{$11.36_{12}$}  &	{$43.9_{7}$}  \\
DMIL-HDP \cite{li2016mlhp}                       &  {$1.34_{14}$}  &	{$4.16_{17}$}  &	{$0.24_{14}$}  &	\cellcolor{bblue}{$0.09_{2}$}  &	{$6.04_{18}$}  &	{$6.60_{15}$}  &	{$9.10_{16}$}  &	{$11.26_{13}$}  &	{$41.2_{14}$}  &  {$1.25_{13}$}  &	{$4.14_{16}$}  &	{$0.23_{17}$}  &	{$0.07_{9}$}  &	{$5.39_{17}$}  &	{$6.99_{11}$}  &	{$8.13_{17}$}  &	{$11.36_{12}$}  &	{$42.5_{15}$}  \\
ULAP \cite{song2018rapid}                        &  {$1.41_{18}$}  &	{$3.97_{11}$}  &	{$0.23_{11}$}  &	\cellcolor{bblue}{$0.09_{2}$}  &	{$5.50_{8}$}  &	\cellcolor{rred}{$5.53_{1}$}  &	{$8.49_{10}$}  &	\cellcolor{bblue}{$10.10_{2}$}  &	\cellcolor{bblue}{$43.1_{2}$}  &  {$1.42_{17}$}  &	{$3.71_{9}$}  &	\cellcolor{rred}{$0.14_{1}$}  &	\cellcolor{yyellow}{$0.06_{3}$}  &	{$4.92_{7}$}  &	\cellcolor{yyellow}{$6.37_{3}$}  &	{$7.31_{4}$}  &	\cellcolor{yyellow}{$10.50_{3}$}  &	\cellcolor{bblue}{$44.5_{2}$}  \\
WaterGAN \cite{li2017watergan}                   &  {$1.25_{6}$}  &	{$3.97_{11}$}  &	{$0.22_{6}$}  &	{$0.10_{11}$}  &	{$5.72_{13}$}  &	{$7.02_{16}$}  &	{$8.91_{14}$}  &	{$11.55_{16}$}  &	{$41.0_{16}$}  &  {$1.14_{6}$}  &	{$3.79_{11}$}  &	\cellcolor{bblue}{$0.18_{2}$}  &	{$0.08_{16}$}  &	{$5.07_{11}$}  &	{$8.02_{18}$}  &	{$7.77_{15}$}  &	{$12.52_{18}$}  &	{$41.9_{17}$}  \\
UGAN \cite{fabbri2018enhancing}                  &  {$1.39_{15}$}  &	{$4.26_{18}$}  &	{$0.25_{17}$}  &	{$0.12_{18}$}  &	{$6.19_{19}$}  &	{$8.22_{19}$}  &	{$9.70_{19}$}  &	{$13.36_{19}$}  &	{$39.0_{19}$}  &  {$1.39_{16}$}  &	{$4.35_{19}$}  &	{$0.26_{19}$}  &	\cellcolor{yyellow}{$0.06_{3}$}  &	{$5.47_{18}$}  &	{$8.32_{19}$}  &	{$8.36_{18}$}  &	{$13.68_{19}$}  &	{$40.3_{19}$}  \\
TUDA \cite{wang2023domain}                       &  {$1.39_{15}$}  &	{$4.01_{13}$}  &	{$0.23_{11}$}  &	\cellcolor{bblue}{$0.09_{2}$}  &	{$6.00_{16}$}  &	{$7.53_{18}$}  &	{$9.03_{15}$}  &	{$12.40_{18}$}  &	{$39.8_{18}$}  &  {$1.10_{5}$}  &	{$4.23_{18}$}  &	{$0.22_{14}$}  &	{$0.07_{9}$}  &	{$5.50_{19}$}  &	{$7.69_{17}$}  &	{$8.44_{19}$}  &	{$12.45_{17}$}  &	{$40.8_{18}$}  \\
TOPAL \cite{jiang2022target}                     &  {$1.22_{4}$}  &	{$3.76_{7}$}  &	\cellcolor{rred}{$0.18_{1}$}  &	{$0.10_{11}$}  &	{$5.49_{6}$}  &	{$6.07_{5}$}  &	{$8.46_{8}$}  &	{$10.53_{7}$}  &	{$43.0_{4}$}  &  {$1.23_{10}$}  &	\cellcolor{rred}{$3.48_{1}$}  &	\cellcolor{bblue}{$0.18_{2}$}  &	\cellcolor{rred}{$0.05_{1}$}  &	{$5.01_{9}$}  &	{$6.88_{9}$}  &	\cellcolor{yyellow}{$7.30_{3}$}  &	{$10.94_{8}$}  &	{$44.1_{4}$}  \\
TACL \cite{liu2022twin}                          &  {$1.29_{12}$}  &	{$3.94_{9}$}  &	{$0.21_{4}$}  &	\cellcolor{rred}{$0.08_{1}$}  &	{$5.77_{14}$}  &	{$7.18_{17}$}  &	{$8.78_{13}$}  &	{$11.77_{17}$}  &	{$40.9_{17}$}  &  {$1.33_{15}$}  &	{$3.67_{8}$}  &	{$0.22_{14}$}  &	\cellcolor{yyellow}{$0.06_{3}$}  &	{$5.14_{13}$}  &	{$7.53_{16}$}  &	{$7.58_{10}$}  &	{$12.06_{16}$}  &	{$42.6_{14}$}  \\
UWCNN \cite{li2020underwater}                       &  {$1.25_{6}$}  &	{$4.15_{16}$}  &	{$0.26_{19}$}  &	{$0.11_{16}$}  &	{$5.53_{9}$}  &	{$6.27_{8}$}  &	{$8.67_{12}$}  &	{$10.71_{9}$}  &	{$41.9_{12}$}  &  {$1.20_{8}$}  &	{$3.94_{13}$}  &	{$0.19_{6}$}  &	{$0.07_{9}$}  &	{$4.88_{5}$}  &	{$6.65_{4}$}  &	{$7.37_{8}$}  &	{$10.94_{8}$}  &	{$43.4_{10}$}  \\
DUIENet \cite{li2019underwater}                  &  {$1.29_{12}$}  &	{$3.71_{5}$}  &	{$0.22_{6}$}  &	\cellcolor{bblue}{$0.09_{2}$}  &	{$5.68_{12}$}  &	{$6.44_{11}$}  &	{$8.46_{8}$}  &	{$10.87_{11}$}  &	{$42.3_{11}$}  &  {$1.29_{14}$}  &	{$4.06_{15}$}  &	{$0.21_{12}$}  &	\cellcolor{rred}{$0.05_{1}$}  &	{$4.93_{8}$}  &	{$6.88_{9}$}  &	{$7.56_{9}$}  &	{$11.18_{11}$}  &	{$43.4_{10}$}  \\
CHE-GLNet \cite{fu2020underwater}                &  {$1.25_{6}$}  &	\cellcolor{rred}{$3.55_{1}$}  &	{$0.22_{6}$}  &	\cellcolor{bblue}{$0.09_{2}$}  &	{$5.66_{11}$}  &	{$6.07_{5}$}  &	{$8.51_{11}$}  &	{$10.31_{4}$}  &	{$42.6_{7}$}  &  {$1.45_{18}$}  &	{$3.64_{7}$}  &	{$0.19_{6}$}  &	{$0.07_{9}$}  &	{$5.06_{10}$}  &	{$7.03_{12}$}  &	{$7.59_{11}$}  &	{$11.14_{10}$}  &	{$43.5_{9}$}  \\
UIEC\textasciicircum{}2Net \cite{wang2021uiec}   &  \cellcolor{yyellow}{$1.20_{3}$}  &	\cellcolor{yyellow}{$3.64_{3}$}  &	{$0.21_{4}$}  &	\cellcolor{bblue}{$0.09_{2}$}  &	{$5.54_{10}$}  &	{$6.42_{9}$}  &	{$8.43_{5}$}  &	{$10.65_{8}$}  &	{$42.9_{5}$}  &  \cellcolor{bblue}{$1.01_{2}$}  &	{$3.56_{4}$}  &	{$0.19_{6}$}  &	{$0.07_{9}$}  &	{$5.15_{14}$}  &	{$6.74_{8}$}  &	{$7.69_{12}$}  &	{$10.85_{6}$}  &	{$43.4_{10}$}  \\
UColor \cite{li2021underwater}                   &  {$1.25_{6}$}  &	\cellcolor{bblue}{$3.62_{2}$}  &	{$0.22_{6}$}  &	\cellcolor{bblue}{$0.09_{2}$}  &	{$5.42_{4}$}  &	{$6.45_{12}$}  &	\cellcolor{bblue}{$8.30_{2}$}  &	{$10.89_{12}$}  &	{$42.4_{10}$}  &  {$1.23_{10}$}  &	{$3.63_{6}$}  &	{$0.22_{14}$}  &	\cellcolor{yyellow}{$0.06_{3}$}  &	{$4.90_{6}$}  &	{$6.69_{6}$}  &	{$7.32_{5}$}  &	{$10.83_{5}$}  &	\cellcolor{yyellow}{$44.2_{3}$}  \\
SGUIE \cite{qi2022sguie}                         &  \cellcolor{bblue}{$1.14_{2}$}  &	{$3.73_{6}$}  &	{$0.24_{14}$}  &	{$0.11_{16}$}  &	\cellcolor{rred}{$5.23_{1}$}  &	{$6.23_{7}$}  &	{$8.36_{4}$}  &	{$10.43_{6}$}  &	\cellcolor{bblue}{$43.1_{2}$}  &  {$1.03_{4}$}  &	{$3.95_{14}$}  &	\cellcolor{bblue}{$0.18_{2}$}  &	{$0.08_{16}$}  &	\cellcolor{yyellow}{$4.80_{3}$}  &	{$6.65_{4}$}  &	{$7.33_{6}$}  &	{$10.76_{4}$}  &	{$44.1_{4}$}  \\
\hline
\hline
& \multicolumn{9}{c||}{\textbf{Faster R-CNN (ResNeXt-101 32$\times$4d, 1x) \cite{ren2015faster}}}                                                     & \multicolumn{9}{c}{\textbf{Faster R-CNN (ResNeXt-101 64$\times$4d, 1x) \cite{ren2015faster}}}   \\ \hline
\textbf{Methods}                               & \textbf{$E_{Cls} \downarrow$}    & \textbf{$E_{Loc} \downarrow$ }   & \textbf{$E_{Both} \downarrow$ }  & \textbf{$E_{Dupe} \downarrow$}   & \textbf{$E_{Bkg} \downarrow$}    & \textbf{$E_{Miss} \downarrow$}   & \textbf{$E_{FP} \downarrow$ }      & \textbf{$E_{FN} \downarrow$ }     & \textbf{AP (\%) $\uparrow$}          & \textbf{$E_{Cls} \downarrow$}     & \textbf{$E_{Loc} \downarrow$ }   & \textbf{$E_{Both} \downarrow$ }   & \textbf{$E_{Dupe} \downarrow$}    & \textbf{$E_{Bkg} \downarrow$}     & \textbf{$E_{Miss} \downarrow$}   & \textbf{$E_{FP} \downarrow$ }     & \textbf{$E_{FN} \downarrow$ }     & \textbf{AP (\%) $\uparrow$}          \\ \hline
RAW                                              &  {$1.24_{8}$}  &	\cellcolor{rred}{$3.32_{1}$}  &	\cellcolor{yyellow}{$0.17_{3}$}  &	{$0.07_{12}$}  &	\cellcolor{yyellow}{$4.68_{3}$}  &	\cellcolor{rred}{$6.21_{1}$}  &	\cellcolor{rred}{$4.97_{1}$}  &	\cellcolor{rred}{$10.07_{1}$}  &	\cellcolor{rred}{$44.6_{1}$}  &  {$1.20_{6}$}  &	\cellcolor{bblue}{$3.31_{2}$}  &	\cellcolor{bblue}{$0.15_{2}$}  &	{$0.06_{6}$}  &	\cellcolor{bblue}{$4.38_{2}$}  &	\cellcolor{rred}{$6.35_{1}$}  &	{$6.54_{4}$}  &	\cellcolor{rred}{$10.43_{1}$}  &	\cellcolor{rred}{$45.3_{1}$}  \\
HE \cite{hummel1977image}                        &  {$1.58_{19}$}  &	{$3.59_{8}$}  &	{$0.23_{17}$}  &	{$0.06_{6}$}  &	{$5.25_{18}$}  &	{$7.74_{15}$}  &	{$7.64_{15}$}  &	{$12.00_{14}$}  &	{$42.4_{15}$}  &  {$1.22_{9}$}  &	{$3.99_{16}$}  &	{$0.20_{14}$}  &	{$0.06_{6}$}  &	{$4.81_{15}$}  &	{$7.67_{12}$}  &	{$7.15_{14}$}  &	{$12.56_{17}$}  &	{$42.7_{14}$}  \\
CLAHE \cite{zuiderveld1994contrast}              &  \cellcolor{bblue}{$1.05_{2}$}  &	{$3.81_{16}$}  &	{$0.18_{4}$}  &	{$0.07_{12}$}  &	{$5.05_{12}$}  &	\cellcolor{yyellow}{$6.64_{3}$}  &	{$7.33_{11}$}  &	{$11.21_{6}$}  &	{$43.4_{12}$}  &  {$1.20_{6}$}  &	{$3.37_{5}$}  &	{$0.22_{17}$}  &	\cellcolor{rred}{$0.05_{1}$}  &	{$4.68_{11}$}  &	\cellcolor{bblue}{$6.57_{2}$}  &	{$6.94_{10}$}  &	\cellcolor{bblue}{$10.65_{2}$}  &	\cellcolor{yyellow}{$44.4_{3}$}  \\
WB \cite{ebner2007color}                         &  {$1.26_{9}$}  &	{$3.78_{13}$}  &	{$0.20_{11}$}  &	{$0.06_{6}$}  &	\cellcolor{yyellow}{$4.68_{3}$}  &	\cellcolor{yyellow}{$6.64_{3}$}  &	{$7.02_{7}$}  &	{$10.96_{4}$}  &	\cellcolor{yyellow}{$44.2_{3}$}  &  {$1.13_{4}$}  &	\cellcolor{rred}{$3.24_{1}$}  &	{$0.17_{8}$}  &	{$0.06_{6}$}  &	{$4.42_{6}$}  &	{$7.68_{13}$}  &	{$6.63_{6}$}  &	{$11.62_{8}$}  &	\cellcolor{yyellow}{$44.4_{3}$}  \\
ACDC \cite{zhang2022underwater}                  &  \cellcolor{rred}{$0.99_{1}$}  &	{$3.58_{7}$}  &	{$0.20_{11}$}  &	{$0.07_{12}$}  &	{$5.24_{17}$}  &	{$7.49_{13}$}  &	{$7.68_{17}$}  &	{$11.89_{13}$}  &	{$42.7_{13}$}  &  \cellcolor{yyellow}{$1.07_{3}$}  &	{$3.72_{13}$}  &	{$0.19_{12}$}  &	{$0.08_{18}$}  &	{$4.99_{18}$}  &	{$7.28_{6}$}  &	{$7.57_{18}$}  &	{$11.69_{11}$}  &	{$43.4_{13}$}  \\
UDCP \cite{drews2016underwater}                  &  {$1.27_{10}$}  &	{$3.45_{4}$}  &	{$0.23_{17}$}  &	\cellcolor{rred}{$0.05_{1}$}  &	{$4.87_{10}$}  &	{$7.26_{10}$}  &	{$7.13_{9}$}  &	{$11.37_{9}$}  &	{$43.8_{9}$}  &  {$1.52_{19}$}  &	{$3.35_{4}$}  &	{$0.22_{17}$}  &	\cellcolor{rred}{$0.05_{1}$}  &	{$4.41_{5}$}  &	{$7.80_{14}$}  &	\cellcolor{rred}{$6.44_{1}$}  &	{$11.69_{11}$}  &	{$44.3_{6}$}  \\
DMIL-HDP \cite{li2016mlhp}                       &  {$1.51_{17}$}  &	{$3.92_{19}$}  &	{$0.21_{13}$}  &	\cellcolor{rred}{$0.05_{1}$}  &	{$4.99_{11}$}  &	{$7.83_{16}$}  &	{$7.24_{10}$}  &	{$12.44_{16}$}  &	{$42.5_{14}$}  &  {$1.21_{8}$}  &	{$4.07_{18}$}  &	{$0.20_{14}$}  &	{$0.06_{6}$}  &	{$4.76_{14}$}  &	{$7.84_{15}$}  &	{$7.23_{16}$}  &	{$12.54_{16}$}  &	{$42.6_{15}$}  \\
ULAP \cite{song2018rapid}                        &  {$1.27_{10}$}  &	\cellcolor{yyellow}{$3.38_{3}$}  &	\cellcolor{bblue}{$0.16_{2}$}  &	\cellcolor{rred}{$0.05_{1}$}  &	{$4.69_{5}$}  &	{$6.68_{5}$}  &	{$6.94_{5}$}  &	\cellcolor{bblue}{$10.81_{2}$}  &	{$44.0_{6}$}  &  {$1.34_{15}$}  &	{$4.06_{17}$}  &	\cellcolor{bblue}{$0.15_{2}$}  &	{$0.06_{6}$}  &	\cellcolor{bblue}{$4.38_{2}$}  &	\cellcolor{yyellow}{$6.69_{3}$}  &	{$6.62_{5}$}  &	{$11.20_{4}$}  &	\cellcolor{bblue}{$44.5_{2}$}  \\
WaterGAN \cite{li2017watergan}                   &  \cellcolor{yyellow}{$1.10_{3}$}  &	{$3.88_{18}$}  &	\cellcolor{rred}{$0.14_{1}$}  &	{$0.07_{12}$}  &	{$4.85_{9}$}  &	{$8.11_{17}$}  &	{$7.35_{13}$}  &	{$12.61_{17}$}  &	{$42.3_{17}$}  &  {$1.24_{11}$}  &	{$3.87_{15}$}  &	\cellcolor{rred}{$0.14_{1}$}  &	{$0.06_{6}$}  &	{$4.70_{12}$}  &	{$8.47_{18}$}  &	{$7.05_{13}$}  &	{$13.05_{18}$}  &	{$42.3_{17}$}  \\
UGAN \cite{fabbri2018enhancing}                  &  {$1.44_{15}$}  &	{$3.78_{13}$}  &	{$0.24_{19}$}  &	{$0.07_{12}$}  &	{$5.18_{16}$}  &	{$8.68_{19}$}  &	{$7.94_{18}$}  &	{$13.36_{19}$}  &	{$40.8_{19}$}  &  {$1.43_{17}$}  &	{$3.83_{14}$}  &	{$0.25_{19}$}  &	{$0.08_{18}$}  &	{$4.89_{17}$}  &	{$9.08_{19}$}  &	{$7.55_{17}$}  &	{$13.74_{19}$}  &	{$41.2_{19}$}  \\
TUDA \cite{wang2023domain}                       &  {$1.39_{13}$}  &	{$3.85_{17}$}  &	{$0.21_{13}$}  &	\cellcolor{rred}{$0.05_{1}$}  &	{$5.58_{19}$}  &	{$8.16_{18}$}  &	{$7.95_{19}$}  &	{$12.67_{18}$}  &	{$41.0_{18}$}  &  \cellcolor{bblue}{$1.02_{2}$}  &	{$3.68_{10}$}  &	{$0.16_{4}$}  &	{$0.07_{15}$}  &	{$5.45_{19}$}  &	{$8.10_{16}$}  &	{$7.87_{19}$}  &	{$12.39_{14}$}  &	{$41.4_{18}$}  \\
TOPAL \cite{jiang2022target}                     &  {$1.49_{16}$}  &	{$3.61_{10}$}  &	{$0.19_{7}$}  &	{$0.06_{6}$}  &	\cellcolor{bblue}{$4.61_{2}$}  &	{$7.07_{7}$}  &	\cellcolor{yyellow}{$6.79_{3}$}  &	{$11.16_{5}$}  &	{$44.1_{4}$}  &  {$1.17_{5}$}  &	\cellcolor{yyellow}{$3.32_{3}$}  &	{$0.19_{12}$}  &	{$0.06_{6}$}  &	{$4.65_{8}$}  &	{$7.51_{9}$}  &	{$6.84_{9}$}  &	{$11.23_{5}$}  &	{$44.1_{8}$}  \\
TACL \cite{liu2022twin}                          &  \cellcolor{yyellow}{$1.10_{3}$}  &	{$3.68_{12}$}  &	{$0.22_{16}$}  &	{$0.07_{12}$}  &	{$5.14_{15}$}  &	{$7.67_{14}$}  &	{$7.67_{16}$}  &	{$12.13_{15}$}  &	{$42.4_{15}$}  &  {$1.40_{16}$}  &	{$3.68_{10}$}  &	{$0.21_{16}$}  &	{$0.07_{15}$}  &	{$4.85_{16}$}  &	{$8.21_{17}$}  &	{$7.18_{15}$}  &	{$12.49_{15}$}  &	{$42.5_{16}$}  \\
UWCNN \cite{li2020underwater}                       &  {$1.19_{6}$}  &	{$3.50_{6}$}  &	{$0.18_{4}$}  &	{$0.07_{12}$}  &	{$4.69_{5}$}  &	{$7.37_{11}$}  &	{$6.88_{4}$}  &	{$11.50_{10}$}  &	{$43.5_{11}$}  &  {$1.29_{13}$}  &	{$3.68_{10}$}  &	{$0.18_{9}$}  &	{$0.06_{6}$}  &	\cellcolor{bblue}{$4.38_{2}$}  &	{$7.56_{10}$}  &	\cellcolor{rred}{$6.44_{1}$}  &	{$11.67_{10}$}  &	{$43.9_{11}$}  \\
DUIENet \cite{li2019underwater}                  &  {$1.51_{17}$}  &	{$3.78_{13}$}  &	{$0.19_{7}$}  &	\cellcolor{rred}{$0.05_{1}$}  &	{$4.84_{8}$}  &	{$7.38_{12}$}  &	{$7.05_{8}$}  &	{$11.51_{11}$}  &	{$43.7_{10}$}  &  {$1.32_{14}$}  &	{$3.63_{9}$}  &	{$0.18_{9}$}  &	\cellcolor{rred}{$0.05_{1}$}  &	{$4.66_{9}$}  &	{$7.18_{5}$}  &	{$6.71_{7}$}  &	{$11.53_{6}$}  &	{$44.2_{7}$}  \\
CHE-GLNet \cite{fu2020underwater}                &  {$1.32_{12}$}  &	\cellcolor{bblue}{$3.37_{2}$}  &	{$0.19_{7}$}  &	{$0.06_{6}$}  &	{$5.05_{12}$}  &	{$7.00_{6}$}  &	{$7.34_{12}$}  &	{$11.22_{7}$}  &	{$43.9_{8}$}  &  \cellcolor{rred}{$0.97_{1}$}  &	{$3.46_{6}$}  &	{$0.16_{4}$}  &	{$0.06_{6}$}  &	{$4.67_{10}$}  &	{$7.50_{8}$}  &	{$6.95_{11}$}  &	{$11.62_{8}$}  &	{$43.8_{12}$}  \\
UIEC\textasciicircum{}2Net \cite{wang2021uiec}   &  {$1.13_{5}$}  &	{$3.45_{4}$}  &	{$0.21_{13}$}  &	{$0.07_{12}$}  &	{$5.07_{14}$}  &	\cellcolor{bblue}{$6.57_{2}$}  &	{$7.42_{14}$}  &	\cellcolor{yyellow}{$10.85_{3}$}  &	{$44.0_{6}$}  &  {$1.25_{12}$}  &	{$3.53_{7}$}  &	{$0.16_{4}$}  &	\cellcolor{rred}{$0.05_{1}$}  &	{$4.61_{7}$}  &	{$7.47_{7}$}  &	{$6.73_{8}$}  &	{$11.58_{7}$}  &	{$44.1_{8}$}  \\
UColor \cite{li2021underwater}                   &  {$1.23_{7}$}  &	{$3.62_{11}$}  &	{$0.19_{7}$}  &	{$0.06_{6}$}  &	{$4.69_{5}$}  &	{$7.07_{7}$}  &	{$6.96_{6}$}  &	{$11.72_{12}$}  &	{$44.1_{4}$}  &  {$1.22_{9}$}  &	{$3.54_{8}$}  &	{$0.16_{4}$}  &	{$0.07_{15}$}  &	{$4.72_{13}$}  &	{$6.98_{4}$}  &	{$6.96_{12}$}  &	\cellcolor{yyellow}{$11.04_{3}$}  &	{$44.1_{8}$}  \\
SGUIE \cite{qi2022sguie}                         &  {$1.41_{14}$}  &	{$3.60_{9}$}  &	{$0.18_{4}$}  &	{$0.06_{6}$}  &	\cellcolor{rred}{$4.50_{1}$}  &	{$7.14_{9}$}  &	\cellcolor{bblue}{$6.75_{2}$}  &	{$11.28_{8}$}  &	\cellcolor{bblue}{$44.3_{2}$}  &  {$1.43_{17}$}  &	{$4.09_{19}$}  &	{$0.18_{9}$}  &	\cellcolor{rred}{$0.05_{1}$}  &	\cellcolor{rred}{$4.22_{1}$}  &	{$7.61_{11}$}  &	\cellcolor{rred}{$6.44_{1}$}  &	{$11.92_{13}$}  &	\cellcolor{yyellow}{$44.4_{3}$}  \\

\hline 
\hline
& \multicolumn{9}{c||}{\textbf{RetinaNet (ResNet-50, 1x) \cite{lin2017focal}}}        & \multicolumn{9}{c}{\textbf{Cascade R-CNN (ResNet-50, 1x) \cite{cai2018cascade}}}      \\ \hline
\textbf{Methods}                               & \textbf{$E_{Cls} \downarrow$}    & \textbf{$E_{Loc} \downarrow$ }   & \textbf{$E_{Both} \downarrow$ }  & \textbf{$E_{Dupe} \downarrow$}   & \textbf{$E_{Bkg} \downarrow$}    & \textbf{$E_{Miss} \downarrow$}   & \textbf{$E_{FP} \downarrow$ }      & \textbf{$E_{FN} \downarrow$ }     & \textbf{AP (\%) $\uparrow$}          & \textbf{$E_{Cls} \downarrow$}     & \textbf{$E_{Loc} \downarrow$ }   & \textbf{$E_{Both} \downarrow$ }   & \textbf{$E_{Dupe} \downarrow$}    & \textbf{$E_{Bkg} \downarrow$}     & \textbf{$E_{Miss} \downarrow$}   & \textbf{$E_{FP} \downarrow$ }     & \textbf{$E_{FN} \downarrow$ }     & \textbf{AP (\%) $\uparrow$}          \\ \hline
RAW                                              &  \cellcolor{bblue}{$1.20_{2}$}  &	\cellcolor{yyellow}{$4.29_{3}$}  &	{$0.18_{6}$}  &	{$0.28_{6}$}  &	{$5.61_{6}$}  &	\cellcolor{bblue}{$3.56_{2}$}  &	\cellcolor{bblue}{$13.10_{2}$}  &	\cellcolor{rred}{$7.15_{1}$}  &	\cellcolor{rred}{$40.7_{1}$}  &  {$1.31_{8}$}  &	{$3.92_{5}$}  &	{$0.23_{18}$}  &	{$0.06_{10}$}  &	\cellcolor{bblue}{$5.51_{2}$}  &	{$6.05_{4}$}  &	\cellcolor{bblue}{$7.95_{2}$}  &	\cellcolor{bblue}{$10.34_{2}$}  &	\cellcolor{rred}{$44.3_{1}$}  \\
HE \cite{hummel1977image}                        &  {$1.30_{8}$}  &	{$4.56_{12}$}  &	{$0.23_{15}$}  &	{$0.31_{15}$}  &	{$6.32_{19}$}  &	{$4.23_{14}$}  &	{$14.72_{15}$}  &	{$8.67_{16}$}  &	{$38.1_{15}$}  &  {$1.37_{15}$}  &	{$4.00_{8}$}  &	{$0.21_{15}$}  &	\cellcolor{yyellow}{$0.05_{3}$}  &	{$6.40_{18}$}  &	{$7.08_{15}$}  &	{$8.88_{16}$}  &	{$12.06_{15}$}  &	{$41.9_{15}$}  \\
CLAHE \cite{zuiderveld1994contrast}              &  {$1.47_{16}$}  &	{$4.40_{6}$}  &	\cellcolor{yyellow}{$0.17_{3}$}  &	\cellcolor{bblue}{$0.25_{2}$}  &	{$5.62_{7}$}  &	{$4.00_{10}$}  &	\cellcolor{yyellow}{$13.25_{3}$}  &	{$7.87_{10}$}  &	{$40.2_{5}$}  &  {$1.32_{11}$}  &	{$3.99_{7}$}  &	{$0.19_{7}$}  &	{$0.07_{15}$}  &	{$5.79_{8}$}  &	\cellcolor{yyellow}{$6.03_{3}$}  &	{$8.38_{9}$}  &	{$10.68_{4}$}  &	{$43.7_{4}$}  \\
WB \cite{ebner2007color}                         &  {$1.43_{15}$}  &	{$4.41_{7}$}  &	\cellcolor{rred}{$0.16_{1}$}  &	{$0.28_{6}$}  &	{$5.52_{4}$}  &	{$3.95_{8}$}  &	{$13.26_{4}$}  &	{$7.86_{8}$}  &	\cellcolor{yyellow}{$40.3_{3}$}  &  {$1.26_{6}$}  &	\cellcolor{rred}{$3.64_{1}$}  &	{$0.19_{7}$}  &	{$0.06_{10}$}  &	\cellcolor{yyellow}{$5.53_{3}$}  &	{$6.36_{7}$}  &	\cellcolor{yyellow}{$8.00_{3}$}  &	\cellcolor{yyellow}{$10.63_{3}$}  &	{$43.7_{4}$}  \\
ACDC \cite{zhang2022underwater}                  &  {$1.31_{9}$}  &	{$4.85_{17}$}  &	{$0.21_{9}$}  &	\cellcolor{rred}{$0.24_{1}$}  &	{$6.23_{15}$}  &	{$4.11_{13}$}  &	{$14.58_{13}$}  &	{$8.26_{13}$}  &	{$38.5_{13}$}  &  \cellcolor{rred}{$1.15_{1}$}  &	{$4.15_{12}$}  &	\cellcolor{bblue}{$0.15_{2}$}  &	\cellcolor{yyellow}{$0.05_{3}$}  &	{$6.37_{16}$}  &	{$6.43_{8}$}  &	{$9.05_{17}$}  &	{$11.53_{12}$}  &	{$42.3_{13}$}  \\
UDCP \cite{drews2016underwater}                  &  {$1.51_{17}$}  &	{$4.52_{11}$}  &	{$0.21_{9}$}  &	{$0.30_{12}$}  &	{$5.68_{8}$}  &	{$3.92_{6}$}  &	{$14.22_{10}$}  &	{$8.04_{12}$}  &	{$39.0_{11}$}  &  {$1.31_{8}$}  &	{$4.10_{10}$}  &	\cellcolor{yyellow}{$0.16_{3}$}  &	\cellcolor{yyellow}{$0.05_{3}$}  &	{$5.67_{4}$}  &	\cellcolor{bblue}{$6.01_{2}$}  &	{$8.10_{5}$}  &	{$11.63_{13}$}  &	{$43.2_{11}$}  \\
DMIL-HDP \cite{li2016mlhp}                       &  {$1.37_{14}$}  &	\cellcolor{rred}{$4.23_{1}$}  &	{$0.25_{17}$}  &	{$0.29_{8}$}  &	{$6.27_{17}$}  &	{$4.07_{12}$}  &	{$14.91_{17}$}  &	{$7.90_{11}$}  &	{$38.5_{13}$}  &  {$1.25_{4}$}  &	{$4.19_{15}$}  &	{$0.20_{11}$}  &	\cellcolor{rred}{$0.04_{1}$}  &	{$6.17_{15}$}  &	{$6.90_{14}$}  &	{$8.59_{14}$}  &	{$12.03_{14}$}  &	{$42.0_{14}$}  \\
ULAP \cite{song2018rapid}                        &  {$1.25_{6}$}  &	{$4.45_{9}$}  &	\cellcolor{yyellow}{$0.17_{3}$}  &	{$0.29_{8}$}  &	{$5.69_{9}$}  &	\cellcolor{yyellow}{$3.80_{3}$}  &	{$13.70_{8}$}  &	\cellcolor{yyellow}{$7.39_{3}$}  &	{$40.0_{6}$}  &  {$1.36_{13}$}  &	{$3.75_{4}$}  &	{$0.18_{5}$}  &	\cellcolor{rred}{$0.04_{1}$}  &	{$5.68_{5}$}  &	\cellcolor{rred}{$5.89_{1}$}  &	{$8.19_{6}$}  &	\cellcolor{rred}{$10.29_{1}$}  &	\cellcolor{bblue}{$44.1_{2}$}  \\
WaterGAN \cite{li2017watergan}                   &  {$1.60_{18}$}  &	{$5.37_{19}$}  &	\cellcolor{rred}{$0.16_{1}$}  &	{$0.32_{18}$}  &	\cellcolor{yyellow}{$5.48_{3}$}  &	{$4.67_{18}$}  &	{$14.77_{16}$}  &	{$9.40_{18}$}  &	{$37.4_{17}$}  &  {$1.38_{16}$}  &	{$4.46_{18}$}  &	\cellcolor{rred}{$0.14_{1}$}  &	\cellcolor{yyellow}{$0.05_{3}$}  &	{$5.74_{7}$}  &	{$7.53_{18}$}  &	{$8.41_{10}$}  &	{$12.48_{18}$}  &	{$41.8_{16}$}  \\
UGAN \cite{fabbri2018enhancing}                  &  {$1.33_{12}$}  &	{$4.91_{18}$}  &	{$0.22_{13}$}  &	{$0.30_{12}$}  &	{$6.28_{18}$}  &	{$4.90_{19}$}  &	{$15.57_{18}$}  &	{$9.52_{19}$}  &	{$36.3_{19}$}  &  {$1.31_{8}$}  &	{$4.49_{19}$}  &	{$0.24_{19}$}  &	{$0.07_{15}$}  &	{$6.42_{19}$}  &	{$8.31_{19}$}  &	{$9.52_{19}$}  &	{$13.76_{19}$}  &	{$39.8_{19}$}  \\
TUDA \cite{wang2023domain}                       &  {$1.62_{19}$}  &	{$4.60_{14}$}  &	{$0.22_{13}$}  &	{$0.31_{15}$}  &	{$6.26_{16}$}  &	{$4.41_{17}$}  &	{$15.78_{19}$}  &	{$8.77_{17}$}  &	{$36.7_{18}$}  &  \cellcolor{bblue}{$1.23_{2}$}  &	{$4.17_{13}$}  &	{$0.21_{15}$}  &	{$0.07_{15}$}  &	{$6.37_{16}$}  &	{$7.42_{16}$}  &	{$9.31_{18}$}  &	{$12.38_{16}$}  &	{$40.6_{18}$}  \\
TOPAL \cite{jiang2022target}                     &  {$1.23_{5}$}  &	{$4.30_{4}$}  &	{$0.21_{9}$}  &	{$0.27_{4}$}  &	\cellcolor{bblue}{$5.37_{2}$}  &	{$3.84_{4}$}  &	\cellcolor{rred}{$13.00_{1}$}  &	{$7.72_{6}$}  &	\cellcolor{bblue}{$40.4_{2}$}  &  {$1.36_{13}$}  &	{$4.21_{16}$}  &	{$0.19_{7}$}  &	\cellcolor{yyellow}{$0.05_{3}$}  &	{$5.85_{11}$}  &	{$6.14_{5}$}  &	{$8.51_{13}$}  &	{$11.00_{5}$}  &	{$43.6_{6}$}  \\
TACL \cite{liu2022twin}                          &  {$1.28_{7}$}  &	{$4.74_{16}$}  &	{$0.25_{17}$}  &	{$0.30_{12}$}  &	{$5.90_{12}$}  &	{$4.32_{15}$}  &	{$14.62_{14}$}  &	{$8.48_{15}$}  &	{$38.1_{15}$}  &  {$1.39_{17}$}  &	{$4.11_{11}$}  &	{$0.20_{11}$}  &	{$0.08_{19}$}  &	{$5.98_{14}$}  &	{$7.43_{17}$}  &	{$8.64_{15}$}  &	{$12.42_{17}$}  &	{$41.7_{17}$}  \\
UWCNN \cite{li2020underwater}                       &  \cellcolor{yyellow}{$1.21_{3}$}  &	{$4.56_{12}$}  &	{$0.20_{7}$}  &	\cellcolor{bblue}{$0.25_{2}$}  &	{$5.58_{5}$}  &	{$4.33_{16}$}  &	{$13.64_{7}$}  &	{$8.38_{14}$}  &	{$38.9_{12}$}  &  {$1.25_{4}$}  &	{$4.24_{17}$}  &	{$0.22_{17}$}  &	{$0.07_{15}$}  &	{$5.79_{8}$}  &	{$6.54_{10}$}  &	{$8.47_{12}$}  &	{$11.46_{11}$}  &	{$42.7_{12}$}  \\
DUIENet \cite{li2019underwater}                  &  \cellcolor{yyellow}{$1.21_{3}$}  &	{$4.31_{5}$}  &	{$0.24_{16}$}  &	{$0.31_{15}$}  &	{$5.78_{11}$}  &	{$4.00_{10}$}  &	{$13.37_{6}$}  &	{$7.76_{7}$}  &	{$39.8_{7}$}  &  {$1.32_{11}$}  &	\cellcolor{bblue}{$3.66_{2}$}  &	{$0.18_{5}$}  &	\cellcolor{yyellow}{$0.05_{3}$}  &	{$5.69_{6}$}  &	{$6.84_{13}$}  &	{$8.07_{4}$}  &	{$11.42_{9}$}  &	{$43.4_{7}$}  \\
CHE-GLNet \cite{fu2020underwater}                &  {$1.32_{11}$}  &	{$4.64_{15}$}  &	{$0.20_{7}$}  &	{$0.29_{8}$}  &	{$6.13_{14}$}  &	\cellcolor{rred}{$3.47_{1}$}  &	{$14.33_{11}$}  &	\cellcolor{bblue}{$7.25_{2}$}  &	{$39.6_{8}$}  &  {$1.46_{18}$}  &	{$3.94_{6}$}  &	{$0.20_{11}$}  &	{$0.06_{10}$}  &	{$5.88_{12}$}  &	{$6.33_{6}$}  &	{$8.43_{11}$}  &	{$11.14_{8}$}  &	{$43.4_{7}$}  \\
UIEC\textasciicircum{}2Net \cite{wang2021uiec}   &  {$1.31_{9}$}  &	\cellcolor{bblue}{$4.27_{2}$}  &	{$0.21_{9}$}  &	{$0.29_{8}$}  &	{$5.90_{12}$}  &	{$3.96_{9}$}  &	{$14.35_{12}$}  &	{$7.48_{4}$}  &	{$39.6_{8}$}  &  {$1.52_{19}$}  &	{$4.18_{14}$}  &	{$0.17_{4}$}  &	{$0.06_{10}$}  &	{$5.89_{13}$}  &	{$6.58_{11}$}  &	{$8.32_{8}$}  &	{$11.42_{9}$}  &	{$43.4_{7}$}  \\
UColor \cite{li2021underwater}                   &  {$1.34_{13}$}  &	{$4.43_{8}$}  &	\cellcolor{yyellow}{$0.17_{3}$}  &	{$0.34_{19}$}  &	{$5.72_{10}$}  &	{$3.84_{4}$}  &	{$13.83_{9}$}  &	{$7.86_{8}$}  &	{$39.5_{10}$}  &  {$1.29_{7}$}  &	\cellcolor{yyellow}{$3.71_{3}$}  &	{$0.19_{7}$}  &	{$0.06_{10}$}  &	{$5.79_{8}$}  &	{$6.47_{9}$}  &	{$8.29_{7}$}  &	{$11.10_{6}$}  &	{$43.4_{7}$}  \\
SGUIE \cite{qi2022sguie}                         &  \cellcolor{rred}{$1.09_{1}$}  &	{$4.49_{10}$}  &	{$0.25_{17}$}  &	{$0.27_{4}$}  &	\cellcolor{rred}{$5.30_{1}$}  &	{$3.92_{6}$}  &	{$13.35_{5}$}  &	{$7.61_{5}$}  &	\cellcolor{yyellow}{$40.3_{3}$}  &  \cellcolor{bblue}{$1.23_{2}$}  &	{$4.07_{9}$}  &	{$0.20_{11}$}  &	\cellcolor{yyellow}{$0.05_{3}$}  &	\cellcolor{rred}{$5.34_{1}$}  &	{$6.58_{11}$}  &	\cellcolor{rred}{$7.70_{1}$}  &	{$11.10_{6}$}  &	\cellcolor{yyellow}{$44.0_{3}$}  \\
\hline
\hline
& \multicolumn{9}{c||}{\textbf{FCOS (ResNet-50, 1x) \cite{tian2019fcos}}}             & \multicolumn{9}{c}{\textbf{ATSS (ResNet-50, 1x) \cite{zhang2020bridging}}}          \\  \hline
\textbf{Methods}                               & \textbf{$E_{Cls} \downarrow$}    & \textbf{$E_{Loc} \downarrow$ }   & \textbf{$E_{Both} \downarrow$ }  & \textbf{$E_{Dupe} \downarrow$}   & \textbf{$E_{Bkg} \downarrow$}    & \textbf{$E_{Miss} \downarrow$}   & \textbf{$E_{FP} \downarrow$ }      & \textbf{$E_{FN} \downarrow$ }     & \textbf{AP (\%) $\uparrow$}          & \textbf{$E_{Cls} \downarrow$}     & \textbf{$E_{Loc} \downarrow$ }   & \textbf{$E_{Both} \downarrow$ }   & \textbf{$E_{Dupe} \downarrow$}    & \textbf{$E_{Bkg} \downarrow$}     & \textbf{$E_{Miss} \downarrow$}   & \textbf{$E_{FP} \downarrow$ }     & \textbf{$E_{FN} \downarrow$ }     & \textbf{AP (\%) $\uparrow$}          \\ \hline
RAW                                              &  {$1.13_{5}$}  &	{$4.13_{7}$}  &	{$0.24_{4}$}  &	{$0.14_{13}$}  &	\cellcolor{yyellow}{$6.96_{3}$}  &	{$8.22_{19}$}  &	\cellcolor{yyellow}{$13.32_{3}$}  &	\cellcolor{rred}{$6.21_{1}$}  &	\cellcolor{rred}{$41.4_{1}$}  &  {$0.98_{5}$}  &	\cellcolor{bblue}{$2.86_{2}$}  &	{$0.29_{4}$}  &	{$0.33_{5}$}  &	{$6.85_{5}$}  &	{$1.67_{4}$}  &	\cellcolor{rred}{$12.41_{1}$}  &	\cellcolor{rred}{$4.21_{1}$}  &	\cellcolor{rred}{$44.8_{1}$}  \\
HE \cite{hummel1977image}                        &  {$1.12_{4}$}  &	{$4.75_{14}$}  &	{$0.24_{4}$}  &	\cellcolor{yyellow}{$0.11_{3}$}  &	{$8.06_{13}$}  &	{$3.21_{14}$}  &	{$15.37_{14}$}  &	{$7.60_{13}$}  &	{$38.1_{11}$}  &  {$1.06_{14}$}  &	{$3.46_{19}$}  &	{$0.29_{4}$}  &	{$0.35_{12}$}  &	{$7.57_{17}$}  &	{$2.05_{16}$}  &	{$14.16_{15}$}  &	{$5.16_{16}$}  &	{$42.2_{14}$}  \\
CLAHE \cite{zuiderveld1994contrast}              &  {$1.17_{8}$}  &	{$4.24_{9}$}  &	{$0.24_{4}$}  &	{$0.14_{13}$}  &	{$7.52_{7}$}  &	{$2.79_{6}$}  &	{$14.02_{6}$}  &	{$6.93_{8}$}  &	{$39.5_{8}$}  &  {$0.99_{7}$}  &	{$3.09_{8}$}  &	\cellcolor{yyellow}{$0.28_{3}$}  &	{$0.33_{5}$}  &	{$6.91_{6}$}  &	{$1.83_{10}$}  &	\cellcolor{yyellow}{$13.02_{3}$}  &	{$4.63_{10}$}  &	\cellcolor{yyellow}{$44.1_{3}$}  \\
WB \cite{ebner2007color}                         &  {$2.02_{18}$}  &	{$5.79_{17}$}  &	\cellcolor{bblue}{$0.21_{2}$}  &	{$0.28_{19}$}  &	{$9.20_{19}$}  &	{$3.99_{16}$}  &	{$19.15_{18}$}  &	{$10.77_{18}$}  &	{$30.5_{18}$}  &  {$1.06_{14}$}  &	{$3.40_{17}$}  &	{$0.32_{9}$}  &	{$0.33_{5}$}  &	\cellcolor{rred}{$6.72_{1}$}  &	\cellcolor{bblue}{$1.65_{2}$}  &	{$13.44_{9}$}  &	{$4.38_{4}$}  &	\cellcolor{yyellow}{$44.1_{3}$}  \\
ACDC \cite{zhang2022underwater}                  &  {$1.20_{9}$}  &	{$5.06_{16}$}  &	{$0.24_{4}$}  &	{$0.14_{13}$}  &	{$8.32_{16}$}  &	{$3.04_{11}$}  &	{$15.66_{15}$}  &	{$7.93_{15}$}  &	{$37.0_{15}$}  &  \cellcolor{bblue}{$0.93_{2}$}  &	{$3.31_{16}$}  &	{$0.30_{6}$}  &	{$0.33_{5}$}  &	{$7.49_{15}$}  &	{$1.76_{7}$}  &	{$14.07_{14}$}  &	{$4.71_{12}$}  &	{$42.7_{13}$}  \\
UDCP \cite{drews2016underwater}                  &  {$1.14_{6}$}  &	{$4.09_{6}$}  &	\cellcolor{rred}{$0.20_{1}$}  &	{$0.12_{7}$}  &	{$7.09_{4}$}  &	\cellcolor{bblue}{$2.70_{2}$}  &	\cellcolor{yyellow}{$13.32_{3}$}  &	{$6.72_{5}$}  &	{$40.4_{5}$}  &  {$1.15_{18}$}  &	{$3.03_{4}$}  &	{$0.34_{13}$}  &	{$0.34_{10}$}  &	{$7.00_{9}$}  &	{$1.67_{4}$}  &	{$13.09_{4}$}  &	{$4.69_{11}$}  &	{$43.6_{7}$}  \\
DMIL-HDP \cite{li2016mlhp}                       &  {$1.22_{13}$}  &	{$4.70_{13}$}  &	{$0.24_{4}$}  &	{$0.13_{8}$}  &	{$8.15_{15}$}  &	{$2.98_{9}$}  &	{$15.27_{13}$}  &	{$7.81_{14}$}  &	{$37.9_{12}$}  &  {$1.00_{8}$}  &	{$3.28_{15}$}  &	{$0.30_{6}$}  &	{$0.38_{18}$}  &	{$7.62_{18}$}  &	{$1.98_{15}$}  &	{$14.34_{16}$}  &	{$4.96_{15}$}  &	{$42.0_{16}$}  \\
ULAP \cite{song2018rapid}                        &  {$1.25_{15}$}  &	{$4.38_{12}$}  &	{$0.27_{12}$}  &	{$0.13_{8}$}  &	{$7.85_{12}$}  &	{$3.17_{13}$}  &	{$15.26_{12}$}  &	{$7.51_{12}$}  &	{$37.5_{14}$}  &  {$0.98_{5}$}  &	{$3.20_{11}$}  &	{$0.31_{8}$}  &	\cellcolor{bblue}{$0.31_{2}$}  &	\cellcolor{yyellow}{$6.79_{3}$}  &	{$1.79_{9}$}  &	\cellcolor{bblue}{$12.94_{2}$}  &	{$4.40_{5}$}  &	{$44.0_{5}$}  \\
WaterGAN \cite{li2017watergan}                   &  {$3.88_{19}$}  &	{$5.85_{19}$}  &	{$0.34_{19}$}  &	{$0.21_{18}$}  &	\cellcolor{rred}{$6.38_{1}$}  &	{$5.20_{18}$}  &	{$25.46_{19}$}  &	{$13.35_{19}$}  &	{$20.8_{19}$}  &  {$1.03_{11}$}  &	{$3.16_{10}$}  &	\cellcolor{bblue}{$0.27_{2}$}  &	{$0.37_{16}$}  &	{$6.97_{7}$}  &	{$8.22_{18}$}  &	{$13.99_{12}$}  &	{$5.26_{17}$}  &	{$42.2_{14}$}  \\
UGAN \cite{fabbri2018enhancing}                  &  {$1.30_{16}$}  &	{$4.78_{15}$}  &	{$0.28_{14}$}  &	{$0.14_{13}$}  &	{$8.49_{17}$}  &	{$3.63_{15}$}  &	{$16.80_{16}$}  &	{$8.41_{16}$}  &	{$36.1_{16}$}  &  {$1.05_{12}$}  &	{$3.24_{12}$}  &	{$0.38_{19}$}  &	{$0.37_{16}$}  &	{$7.75_{19}$}  &	{$8.22_{18}$}  &	{$15.99_{19}$}  &	{$5.48_{18}$}  &	{$40.5_{19}$}  \\
TUDA \cite{wang2023domain}                       &  {$1.24_{14}$}  &	{$4.00_{5}$}  &	{$0.30_{17}$}  &	\cellcolor{yyellow}{$0.11_{3}$}  &	{$8.13_{14}$}  &	{$3.09_{12}$}  &	{$15.08_{11}$}  &	{$7.49_{11}$}  &	{$37.8_{13}$}  &  {$1.08_{16}$}  &	{$3.40_{17}$}  &	{$0.32_{9}$}  &	{$0.34_{10}$}  &	{$7.45_{14}$}  &	{$2.18_{17}$}  &	{$14.40_{17}$}  &	{$5.65_{19}$}  &	{$41.1_{18}$}  \\
TOPAL \cite{jiang2022target}                     &  \cellcolor{rred}{$1.07_{1}$}  &	\cellcolor{yyellow}{$3.79_{3}$}  &	{$0.26_{10}$}  &	{$0.13_{8}$}  &	\cellcolor{bblue}{$6.94_{2}$}  &	\cellcolor{rred}{$2.66_{1}$}  &	\cellcolor{bblue}{$13.17_{2}$}  &	\cellcolor{yyellow}{$6.52_{3}$}  &	\cellcolor{bblue}{$40.7_{2}$}  &  \cellcolor{rred}{$0.91_{1}$}  &	{$3.08_{6}$}  &	\cellcolor{rred}{$0.25_{1}$}  &	\cellcolor{bblue}{$0.31_{2}$}  &	{$6.99_{8}$}  &	{$1.68_{6}$}  &	{$13.41_{8}$}  &	{$4.52_{6}$}  &	\cellcolor{bblue}{$44.2_{2}$}  \\
TACL \cite{liu2022twin}                          &  {$1.14_{6}$}  &	{$4.28_{10}$}  &	{$0.30_{17}$}  &	{$0.13_{8}$}  &	{$7.62_{9}$}  &	{$2.91_{8}$}  &	{$14.70_{10}$}  &	{$7.06_{10}$}  &	{$38.9_{10}$}  &  {$1.17_{19}$}  &	{$3.27_{14}$}  &	{$0.33_{11}$}  &	{$0.36_{14}$}  &	{$7.54_{16}$}  &	\cellcolor{rred}{$1.64_{1}$}  &	{$14.83_{18}$}  &	{$4.71_{12}$}  &	{$41.9_{17}$}  \\
UWCNN \cite{li2020underwater}                       &  {$1.53_{17}$}  &	{$5.84_{18}$}  &	\cellcolor{bblue}{$0.21_{2}$}  &	{$0.19_{17}$}  &	{$8.82_{18}$}  &	{$4.27_{17}$}  &	{$19.02_{17}$}  &	{$10.25_{17}$}  &	{$31.2_{17}$}  &  \cellcolor{yyellow}{$0.96_{3}$}  &	{$3.08_{6}$}  &	{$0.36_{17}$}  &	\cellcolor{rred}{$0.30_{1}$}  &	{$6.80_{4}$}  &	{$1.89_{14}$}  &	{$13.21_{5}$}  &	{$4.81_{14}$}  &	{$42.9_{12}$}  \\
DUIENet \cite{li2019underwater}                  &  {$1.20_{9}$}  &	{$4.16_{8}$}  &	{$0.25_{9}$}  &	\cellcolor{rred}{$0.10_{1}$}  &	{$7.59_{8}$}  &	{$2.78_{5}$}  &	{$14.03_{7}$}  &	{$6.87_{6}$}  &	{$39.5_{8}$}  &  {$1.02_{10}$}  &	{$3.10_{9}$}  &	{$0.35_{15}$}  &	{$0.33_{5}$}  &	{$7.00_{9}$}  &	{$1.78_{8}$}  &	{$13.70_{11}$}  &	{$4.57_{8}$}  &	{$43.2_{11}$}  \\
CHE-GLNet \cite{fu2020underwater}                &  {$1.20_{9}$}  &	{$4.37_{11}$}  &	{$0.28_{14}$}  &	\cellcolor{yyellow}{$0.11_{3}$}  &	{$7.73_{10}$}  &	{$2.77_{4}$}  &	{$14.35_{9}$}  &	{$6.91_{7}$}  &	{$40.0_{7}$}  &  {$1.05_{12}$}  &	{$3.25_{13}$}  &	{$0.33_{11}$}  &	{$0.32_{4}$}  &	{$7.41_{13}$}  &	{$1.86_{12}$}  &	{$14.03_{13}$}  &	{$4.56_{7}$}  &	{$43.4_{10}$}  \\
UIEC\textasciicircum{}2Net \cite{wang2021uiec}   &  {$1.20_{9}$}  &	{$3.97_{4}$}  &	{$0.27_{12}$}  &	\cellcolor{rred}{$0.10_{1}$}  &	{$7.84_{11}$}  &	{$2.98_{9}$}  &	{$14.19_{8}$}  &	{$7.00_{9}$}  &	{$40.1_{6}$}  &  {$1.13_{17}$}  &	{$3.04_{5}$}  &	{$0.35_{15}$}  &	{$0.35_{12}$}  &	{$7.12_{11}$}  &	{$1.87_{13}$}  &	{$13.39_{7}$}  &	{$4.58_{9}$}  &	{$43.6_{7}$}  \\
UColor \cite{li2021underwater}                   &  \cellcolor{bblue}{$1.09_{2}$}  &	\cellcolor{rred}{$3.69_{1}$}  &	{$0.26_{10}$}  &	{$0.13_{8}$}  &	{$7.13_{5}$}  &	{$2.82_{7}$}  &	{$13.36_{5}$}  &	\cellcolor{bblue}{$6.48_{2}$}  &	{$40.5_{4}$}  &  {$1.00_{8}$}  &	\cellcolor{yyellow}{$3.01_{3}$}  &	{$0.34_{13}$}  &	{$0.39_{19}$}  &	{$7.14_{12}$}  &	\cellcolor{bblue}{$1.65_{2}$}  &	{$13.63_{10}$}  &	\cellcolor{bblue}{$4.26_{2}$}  &	{$43.6_{7}$}  \\
SGUIE \cite{qi2022sguie}                         &  \cellcolor{yyellow}{$1.11_{3}$}  &	\cellcolor{bblue}{$3.73_{2}$}  &	{$0.28_{14}$}  &	\cellcolor{yyellow}{$0.11_{3}$}  &	{$7.16_{6}$}  &	\cellcolor{yyellow}{$2.72_{3}$}  &	\cellcolor{rred}{$13.10_{1}$}  &	{$6.55_{4}$}  &	\cellcolor{bblue}{$40.7_{2}$}  &  \cellcolor{yyellow}{$0.96_{3}$}  &	\cellcolor{rred}{$2.85_{1}$}  &	{$0.36_{17}$}  &	{$0.36_{14}$}  &	\cellcolor{bblue}{$6.74_{2}$}  &	{$1.84_{11}$}  &	{$13.28_{6}$}  &	\cellcolor{yyellow}{$4.37_{3}$}  &	{$43.9_{6}$}  \\
\hline 
\hline
& \multicolumn{9}{c||}{\textbf{TOOD (ResNet-50, 1x) \cite{feng2021tood}}}             & \multicolumn{9}{c}{\textbf{SSD (VGG16, 120e) \cite{liu2016ssd}}}                    \\ \hline
\textbf{Methods}                               & \textbf{$E_{Cls} \downarrow$}    & \textbf{$E_{Loc} \downarrow$ }   & \textbf{$E_{Both} \downarrow$ }  & \textbf{$E_{Dupe} \downarrow$}   & \textbf{$E_{Bkg} \downarrow$}    & \textbf{$E_{Miss} \downarrow$}   & \textbf{$E_{FP} \downarrow$ }      & \textbf{$E_{FN} \downarrow$ }     & \textbf{AP (\%) $\uparrow$}          & \textbf{$E_{Cls} \downarrow$}     & \textbf{$E_{Loc} \downarrow$ }   & \textbf{$E_{Both} \downarrow$ }   & \textbf{$E_{Dupe} \downarrow$}    & \textbf{$E_{Bkg} \downarrow$}     & \textbf{$E_{Miss} \downarrow$}   & \textbf{$E_{FP} \downarrow$ }     & \textbf{$E_{FN} \downarrow$ }     & \textbf{AP (\%) $\uparrow$}          \\ \hline
RAW                                              &  \cellcolor{yyellow}{$0.84_{3}$}  &	\cellcolor{rred}{$3.02_{1}$}  &	{$0.30_{10}$}  &	{$0.66_{19}$}  &	{$6.59_{4}$}  &	\cellcolor{rred}{$1.71_{1}$}  &	{$13.32_{7}$}  &	\cellcolor{rred}{$3.39_{1}$}  &	\cellcolor{rred}{$45.4_{1}$}  &  \cellcolor{yyellow}{$1.31_{3}$}  &	\cellcolor{yyellow}{$5.42_{3}$}  &	\cellcolor{yyellow}{$0.22_{3}$}  &	{$0.08_{8}$}  &	\cellcolor{yyellow}{$8.77_{3}$}  &	{$2.92_{9}$}  &	\cellcolor{bblue}{$16.25_{2}$}  &	{$9.15_{6}$}  &	\cellcolor{rred}{$35.1_{1}$}  \\
HE \cite{hummel1977image}                        &  {$1.00_{14}$}  &	{$3.20_{4}$}  &	\cellcolor{rred}{$0.20_{1}$}  &	\cellcolor{bblue}{$0.54_{2}$}  &	{$7.79_{17}$}  &	{$2.03_{14}$}  &	{$14.82_{16}$}  &	{$4.84_{9}$}  &	{$42.5_{16}$}  &  {$1.48_{18}$}  &	{$5.88_{13}$}  &	{$0.25_{7}$}  &	\cellcolor{yyellow}{$0.07_{3}$}  &	{$9.89_{19}$}  &	{$3.38_{16}$}  &	{$18.07_{16}$}  &	{$10.49_{16}$}  &	{$32.6_{15}$}  \\
CLAHE \cite{zuiderveld1994contrast}              &  {$0.98_{12}$}  &	{$3.22_{6}$}  &	{$0.29_{7}$}  &	{$0.60_{11}$}  &	{$6.75_{5}$}  &	\cellcolor{bblue}{$1.77_{2}$}  &	{$13.29_{4}$}  &	\cellcolor{yyellow}{$4.47_{3}$}  &	\cellcolor{yyellow}{$44.7_{3}$}  &  {$1.37_{8}$}  &	\cellcolor{bblue}{$5.28_{2}$}  &	\cellcolor{rred}{$0.20_{1}$}  &	{$0.08_{8}$}  &	{$9.01_{10}$}  &	{$2.89_{8}$}  &	{$16.45_{6}$}  &	{$9.14_{5}$}  &	\cellcolor{yyellow}{$35.0_{3}$}  \\
WB \cite{ebner2007color}                         &  {$0.85_{4}$}  &	{$3.31_{7}$}  &	{$0.31_{13}$}  &	{$0.63_{15}$}  &	{$6.85_{8}$}  &	{$1.84_{6}$}  &	{$13.65_{10}$}  &	{$4.57_{4}$}  &	{$44.4_{4}$}  &  \cellcolor{yyellow}{$1.31_{3}$}  &	{$6.13_{16}$}  &	\cellcolor{yyellow}{$0.22_{3}$}  &	\cellcolor{rred}{$0.06_{1}$}  &	{$8.89_{6}$}  &	{$3.03_{11}$}  &	{$16.55_{7}$}  &	{$9.66_{12}$}  &	{$34.3_{11}$}  \\
ACDC \cite{zhang2022underwater}                  &  \cellcolor{bblue}{$0.83_{2}$}  &	{$3.32_{9}$}  &	{$0.30_{10}$}  &	{$0.55_{5}$}  &	{$7.28_{14}$}  &	{$1.98_{10}$}  &	{$14.44_{13}$}  &	{$4.82_{8}$}  &	{$43.2_{13}$}  &  {$1.40_{11}$}  &	{$5.50_{7}$}  &	\cellcolor{bblue}{$0.21_{2}$}  &	{$0.10_{16}$}  &	{$9.46_{16}$}  &	{$3.13_{13}$}  &	{$17.26_{13}$}  &	{$9.17_{8}$}  &	{$33.9_{12}$}  \\
UDCP \cite{drews2016underwater}                  &  {$0.97_{11}$}  &	{$3.48_{16}$}  &	{$0.29_{7}$}  &	{$0.56_{6}$}  &	{$6.82_{7}$}  &	{$1.80_{4}$}  &	{$13.30_{6}$}  &	{$4.84_{9}$}  &	{$44.3_{7}$}  &  {$1.39_{10}$}  &	{$5.54_{8}$}  &	{$0.23_{5}$}  &	{$0.08_{8}$}  &	{$9.06_{12}$}  &	{$3.04_{12}$}  &	{$16.87_{11}$}  &	{$9.23_{9}$}  &	{$34.4_{9}$}  \\
DMIL-HDP \cite{li2016mlhp}                       &  {$0.88_{7}$}  &	{$3.39_{13}$}  &	{$0.28_{4}$}  &	\cellcolor{bblue}{$0.54_{2}$}  &	{$7.85_{18}$}  &	{$1.98_{10}$}  &	{$14.74_{14}$}  &	{$4.96_{14}$}  &	{$42.6_{15}$}  &  {$1.43_{15}$}  &	{$5.75_{12}$}  &	{$0.32_{19}$}  &	{$0.08_{8}$}  &	{$9.83_{17}$}  &	{$3.35_{15}$}  &	{$18.22_{17}$}  &	{$10.16_{15}$}  &	{$32.6_{15}$}  \\
ULAP \cite{song2018rapid}                        &  {$0.92_{9}$}  &	\cellcolor{bblue}{$3.07_{2}$}  &	{$0.28_{4}$}  &	{$0.58_{10}$}  &	\cellcolor{yyellow}{$6.54_{3}$}  &	{$1.86_{7}$}  &	\cellcolor{bblue}{$12.94_{2}$}  &	\cellcolor{bblue}{$4.40_{2}$}  &	\cellcolor{bblue}{$44.8_{2}$}  &  {$1.47_{17}$}  &	{$5.61_{10}$}  &	{$0.27_{13}$}  &	{$0.08_{8}$}  &	{$8.81_{5}$}  &	{$3.17_{14}$}  &	{$16.41_{4}$}  &	{$9.75_{13}$}  &	{$34.4_{9}$}  \\
WaterGAN \cite{li2017watergan}                   &  {$0.94_{10}$}  &	{$3.38_{11}$}  &	\cellcolor{bblue}{$0.25_{2}$}  &	{$0.61_{12}$}  &	{$7.05_{11}$}  &	{$2.18_{17}$}  &	{$14.34_{12}$}  &	{$5.48_{17}$}  &	{$42.4_{17}$}  &  {$1.42_{12}$}  &	{$6.56_{19}$}  &	{$0.26_{9}$}  &	\cellcolor{yyellow}{$0.07_{3}$}  &	{$8.89_{6}$}  &	{$3.95_{18}$}  &	{$17.32_{14}$}  &	{$12.08_{18}$}  &	{$31.5_{17}$}  \\
UGAN \cite{fabbri2018enhancing}                  &  {$1.04_{17}$}  &	{$3.45_{15}$}  &	{$0.35_{17}$}  &	{$0.65_{18}$}  &	{$7.64_{15}$}  &	{$2.44_{18}$}  &	{$16.37_{19}$}  &	{$5.62_{18}$}  &	{$40.8_{19}$}  &  {$1.51_{19}$}  &	{$6.55_{18}$}  &	{$0.28_{15}$}  &	\cellcolor{rred}{$0.06_{1}$}  &	{$9.38_{15}$}  &	{$4.24_{19}$}  &	{$18.91_{19}$}  &	{$12.35_{19}$}  &	{$30.1_{19}$}  \\
TUDA \cite{wang2023domain}                       &  {$1.04_{17}$}  &	{$3.58_{17}$}  &	{$0.36_{18}$}  &	{$0.57_{7}$}  &	{$7.76_{16}$}  &	{$1.98_{10}$}  &	{$15.73_{17}$}  &	{$5.36_{16}$}  &	{$41.1_{18}$}  &  {$1.42_{12}$}  &	{$6.52_{17}$}  &	{$0.26_{9}$}  &	\cellcolor{yyellow}{$0.07_{3}$}  &	{$9.86_{18}$}  &	{$3.83_{17}$}  &	{$18.85_{18}$}  &	{$11.87_{17}$}  &	{$30.6_{18}$}  \\
TOPAL \cite{jiang2022target}                     &  {$0.85_{4}$}  &	{$3.31_{7}$}  &	{$0.28_{4}$}  &	{$0.64_{17}$}  &	{$6.75_{5}$}  &	{$1.82_{5}$}  &	{$13.64_{9}$}  &	{$4.57_{4}$}  &	{$44.4_{4}$}  &  {$1.32_{5}$}  &	{$5.48_{6}$}  &	{$0.24_{6}$}  &	{$0.09_{14}$}  &	\cellcolor{bblue}{$8.72_{2}$}  &	{$2.77_{6}$}  &	{$16.42_{5}$}  &	\cellcolor{bblue}{$8.85_{2}$}  &	{$34.9_{5}$}  \\
TACL \cite{liu2022twin}                          &  {$0.88_{7}$}  &	{$3.61_{18}$}  &	{$0.33_{15}$}  &	{$0.63_{15}$}  &	{$7.23_{13}$}  &	{$1.93_{8}$}  &	{$14.75_{15}$}  &	{$4.88_{11}$}  &	{$42.7_{14}$}  &  {$1.38_{9}$}  &	{$6.06_{15}$}  &	{$0.31_{16}$}  &	{$0.11_{18}$}  &	{$9.13_{13}$}  &	{$2.98_{10}$}  &	{$17.45_{15}$}  &	{$9.78_{14}$}  &	{$33.0_{14}$}  \\
UWCNN \cite{li2020underwater}                       &  \cellcolor{rred}{$0.82_{1}$}  &	\cellcolor{yyellow}{$3.15_{3}$}  &	{$0.31_{13}$}  &	{$0.62_{14}$}  &	\cellcolor{rred}{$6.48_{1}$}  &	{$2.10_{15}$}  &	\cellcolor{rred}{$12.88_{1}$}  &	{$5.06_{15}$}  &	{$44.0_{11}$}  &  {$1.34_{6}$}  &	{$6.04_{14}$}  &	{$0.27_{13}$}  &	{$0.10_{16}$}  &	{$9.01_{10}$}  &	{$2.76_{5}$}  &	{$17.12_{12}$}  &	{$9.50_{11}$}  &	{$33.5_{13}$}  \\
DUIENet \cite{li2019underwater}                  &  {$1.03_{16}$}  &	{$3.21_{5}$}  &	{$0.36_{18}$}  &	{$0.57_{7}$}  &	{$6.89_{9}$}  &	{$2.10_{15}$}  &	{$13.77_{11}$}  &	{$4.89_{12}$}  &	{$43.9_{12}$}  &  {$1.42_{12}$}  &	{$5.68_{11}$}  &	{$0.25_{7}$}  &	\cellcolor{yyellow}{$0.07_{3}$}  &	{$8.95_{9}$}  &	{$2.84_{7}$}  &	{$16.59_{8}$}  &	{$9.30_{10}$}  &	{$34.8_{6}$}  \\
CHE-GLNet \cite{fu2020underwater}                &  {$0.99_{13}$}  &	{$3.38_{11}$}  &	{$0.29_{7}$}  &	{$0.57_{7}$}  &	{$6.90_{10}$}  &	{$1.94_{9}$}  &	{$13.29_{4}$}  &	{$4.91_{13}$}  &	{$44.1_{10}$}  &  {$1.34_{6}$}  &	{$5.43_{4}$}  &	{$0.26_{9}$}  &	{$0.08_{8}$}  &	{$9.14_{14}$}  &	\cellcolor{bblue}{$2.60_{2}$}  &	{$16.83_{10}$}  &	{$9.04_{4}$}  &	{$34.7_{8}$}  \\
UIEC\textasciicircum{}2Net \cite{wang2021uiec}   &  {$1.02_{15}$}  &	{$3.36_{10}$}  &	{$0.30_{10}$}  &	\cellcolor{bblue}{$0.54_{2}$}  &	{$7.12_{12}$}  &	\cellcolor{bblue}{$1.77_{2}$}  &	{$13.60_{8}$}  &	{$4.63_{6}$}  &	{$44.3_{7}$}  &  {$1.45_{16}$}  &	{$5.43_{4}$}  &	{$0.31_{16}$}  &	\cellcolor{yyellow}{$0.07_{3}$}  &	{$8.89_{6}$}  &	\cellcolor{yyellow}{$2.63_{3}$}  &	{$16.66_{9}$}  &	\cellcolor{yyellow}{$8.86_{3}$}  &	{$34.8_{6}$}  \\
UColor \cite{li2021underwater}                   &  {$1.27_{19}$}  &	{$5.60_{19}$}  &	\cellcolor{yyellow}{$0.26_{3}$}  &	\cellcolor{rred}{$0.09_{1}$}  &	{$8.49_{19}$}  &	{$2.71_{19}$}  &	{$15.77_{18}$}  &	{$9.19_{19}$}  &	{$44.3_{7}$}  &  \cellcolor{rred}{$0.93_{1}$}  &	\cellcolor{rred}{$3.00_{1}$}  &	{$0.31_{16}$}  &	{$0.58_{19}$}  &	\cellcolor{rred}{$6.92_{1}$}  &	\cellcolor{rred}{$1.89_{1}$}  &	\cellcolor{rred}{$13.29_{1}$}  &	\cellcolor{rred}{$4.55_{1}$}  &	\cellcolor{rred}{$35.1_{1}$}  \\
SGUIE \cite{qi2022sguie}                         &  {$0.85_{4}$}  &	{$3.41_{14}$}  &	{$0.34_{16}$}  &	{$0.61_{12}$}  &	\cellcolor{bblue}{$6.53_{2}$}  &	{$1.99_{13}$}  &	\cellcolor{yyellow}{$13.20_{3}$}  &	{$4.73_{7}$}  &	{$44.4_{4}$}  &  \cellcolor{bblue}{$1.20_{2}$}  &	{$5.56_{9}$}  &	{$0.26_{9}$}  &	{$0.09_{14}$}  &	{$8.79_{4}$}  &	{$2.68_{4}$}  &	\cellcolor{yyellow}{$16.26_{3}$}  &	{$9.15_{6}$}  &	\cellcolor{yyellow}{$35.0_{3}$}  \\
\hline
\hline
\end{tabular}}
\label{table5}
\end{table*}

\begin{figure*}[!t]    
  \centering
  \includegraphics[width=0.95\linewidth]{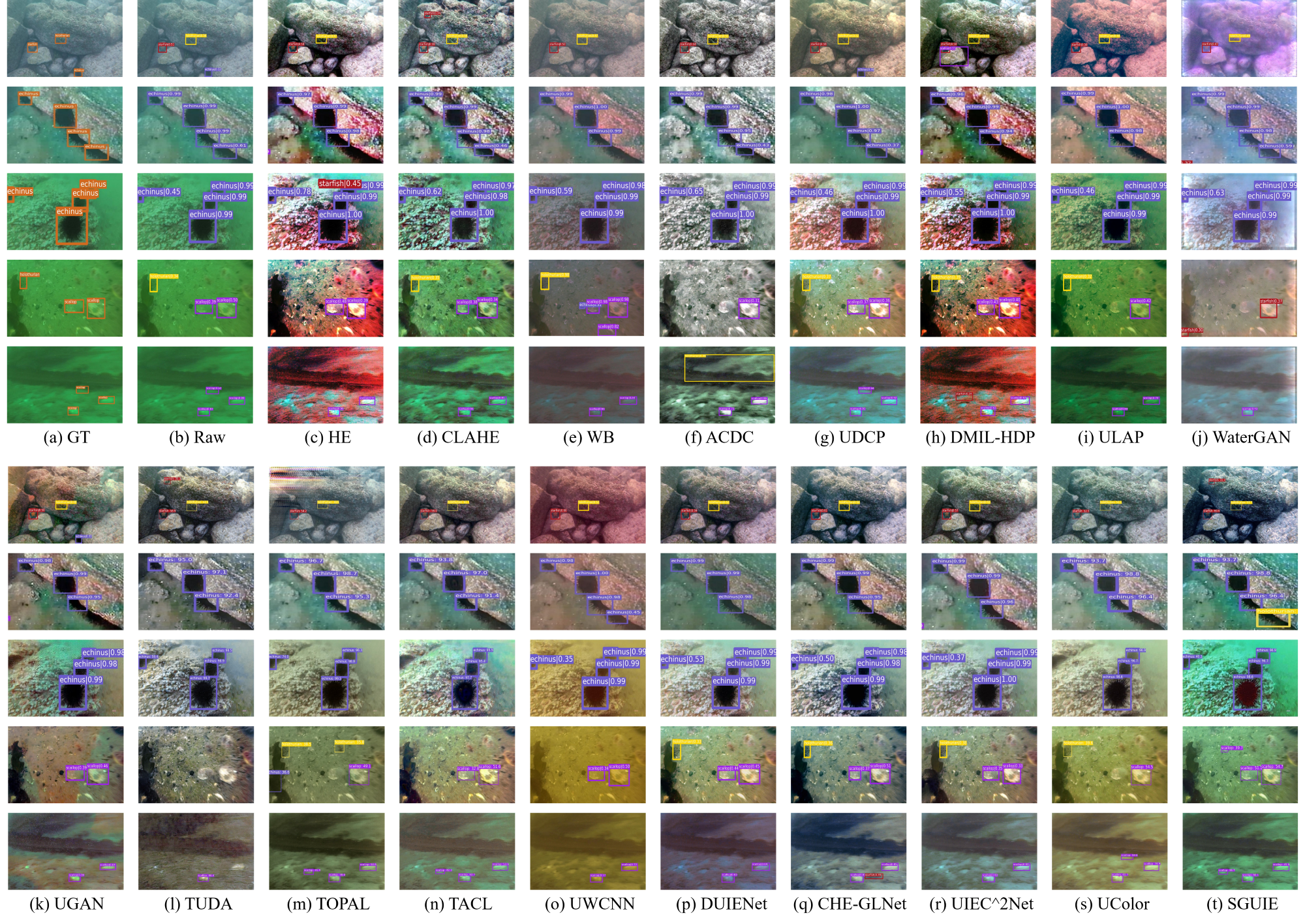}
  \caption{\textbf{Samples of the visual detection results.} From top to bottom are ATSS \cite{zhang2020bridging}, Cascade R-CNN \cite{cai2018cascade}, Faster R-CNN \cite{ren2015faster}, FCOS \cite{tian2019fcos}, and SSD \cite{liu2016ssd} that are trained and tested on (b) raw and results enhanced using (c) HE \cite{hummel1977image}, (d) CLAHE \cite{zuiderveld1994contrast}, (e) WB \cite{ebner2007color}, (f) ACDC \cite{zhang2022underwater}, (g) UDCP \cite{drews2016underwater}, (h) DMIL-HDP \cite{li2016mlhp}, (i) ULAP \cite{song2018rapid}, (j) WaterGAN \cite{li2017watergan}, (k) UGAN \cite{fabbri2018enhancing}, (l) TUDA \cite{wang2023domain}, (m) TOPAL \cite{jiang2022target}, (n) TACL \cite{liu2022twin}, (o) UWCNN \cite{li2020underwater}, (p) DUIENet \cite{li2019underwater}, (q) CHE-GLNet \cite{fu2020underwater}, (r) UIEC\^{}2Net \cite{wang2021uiec}, (s) UColor \cite{li2021underwater}, and (t) SGUIE \cite{qi2022sguie}. Note that the bounding boxes with different colors represent different categories, the \textcolor[RGB]{210,105,30}{orange}, \textcolor{yyellow}{yellow}, \textcolor[RGB]{106,90,205}{purple}, \textcolor[RGB]{160,32,240}{light-purple}, and \textcolor[RGB]{176,23,31}{red} bounding box means GT, holothurian, echinus, scallop, and starfish, respectively. (Zoom in for best view.)}
  \label{fig4}
\end{figure*}

\textbf{TIDE} (\textbf{T}oolbox for \textbf{I}dentifying Object \textbf{D}etection Errors) \cite{tideeccv2020} is a framework and associated toolbox for analyzing the source of error in object detection methods.
TIDE segments errors into six types, including 
1) classification error ($E_{Cls}$), which means localized correctly but classified incorrectly; 
2) localization error ($E_{Loc}$), which means classified correctly but localized incorrectly; 
3) both Cls and Loc error ($E_{Both}$), which means both classified and localized incorrectly; 
4) duplicate detection error ($E_{Dupe}$), which means GT has match another higher-scoring detection bounding box;
5) background error ($E_{Bkg}$), which means detected background as an object; and
6) Missed GT error ($E_{Miss}$), which means all undetected ground truth (false negatives) not already covered by classification or localization error.
Meanwhile, TIDE also provides two special error types (False Positive error ($E_{FP}$) and False Negative error ($E_{FN}$)). 

To better answer the question: \textit{``How does underwater image enhancement contribute to object detection?''} We perform an error analysis using TIDE to study how the different underwater image enhancement algorithms affect object detection. 
Table \ref{table5} presents the error scores of Faster R-CNN \cite{ren2015faster} with different backbones and training schedules, RetinaNet \cite{lin2017focal}, Cascade R-CNN \cite{cai2018cascade}, FCOS \cite{tian2019fcos}, ATSS \cite{zhang2020bridging}, TOOD\cite{feng2021tood}, and SSD \cite{liu2016ssd}.
We first comprehensively compare the errors between different detectors. 
We find that classification, localization, and missed GT errors in ATSS \cite{zhang2020bridging} and TOOD \cite{feng2021tood} are lower than other detectors, while the background errors are higher than two-stage detectors (Faster R-CNN \cite{ren2015faster} and Cascade R-CNN \cite{cai2018cascade}). 
Both Cls and Loc error and duplicate detection error have little difference in each detector. %
Meanwhile, with the enhancement of the detector backbone, the background error becomes smaller.
Besides, the classification error and localization error of raw domain Faster R-CNN \cite{ren2015faster} are higher than most of other domain detectors. 
However, in RetinaNet \cite{lin2017focal}, the raw domain detector has the second lowest classification error and the third lowest error in localization. The percentage of both error and duplication detection error in each detector is minimal, and the fluctuation can be negligible. 
The background error and miss GT error have the greatest impact on the AP values. 
With an increase in background error or miss GT error, the AP value decreases.
For example, UGAN domain detectors, including Faster R-CNN \cite{ren2015faster}, RetinaNet \cite{lin2017focal}, Cascade R-CNN \cite{cai2018cascade}, ATSS \cite{zhang2020bridging}, TOOD \cite{feng2021tood}, and SSD \cite{liu2016ssd} \cite{fabbri2018enhancing}, have the highest background \& miss GT error and achieve the lowest detection performance.
This phenomenon illustrates that  \textit{underwater image enhancement significantly increases the probability that the detector detects the background as foreground or cannot fully detect the accurate object.}
One more example is that the background error and miss GT error of raw domain detection in Faster R-CNN is 5.27 \& 5.67, while ULAP \cite{song2018rapid} domain detection is 5.50 \& 5.53, respectively.
The ULAP has a lower miss GT error, which means ULAP can reduce the probability of miss detection.
However, other errors in ULAP are higher than the raw-domain. This eventually leads to the fact that the AP value of ULAP domain detector is lower than the raw domain. Similar situations happen in RetinaNet \cite{lin2017focal} and ATSS \cite{zhang2020bridging}, in which the WB \cite{ebner2007color} domain detection has a lower background error than raw domain detection, but other errors are higher than the raw domain. Thus, the AP value of WB domain detection is lower than the raw domain. Meanwhile, we also find that most raw domain detectors have the lowest error value on two special error types (FP error and FN error).

\begin{figure*}[!t]    
  \centering
  \includegraphics[width=1.0\linewidth]{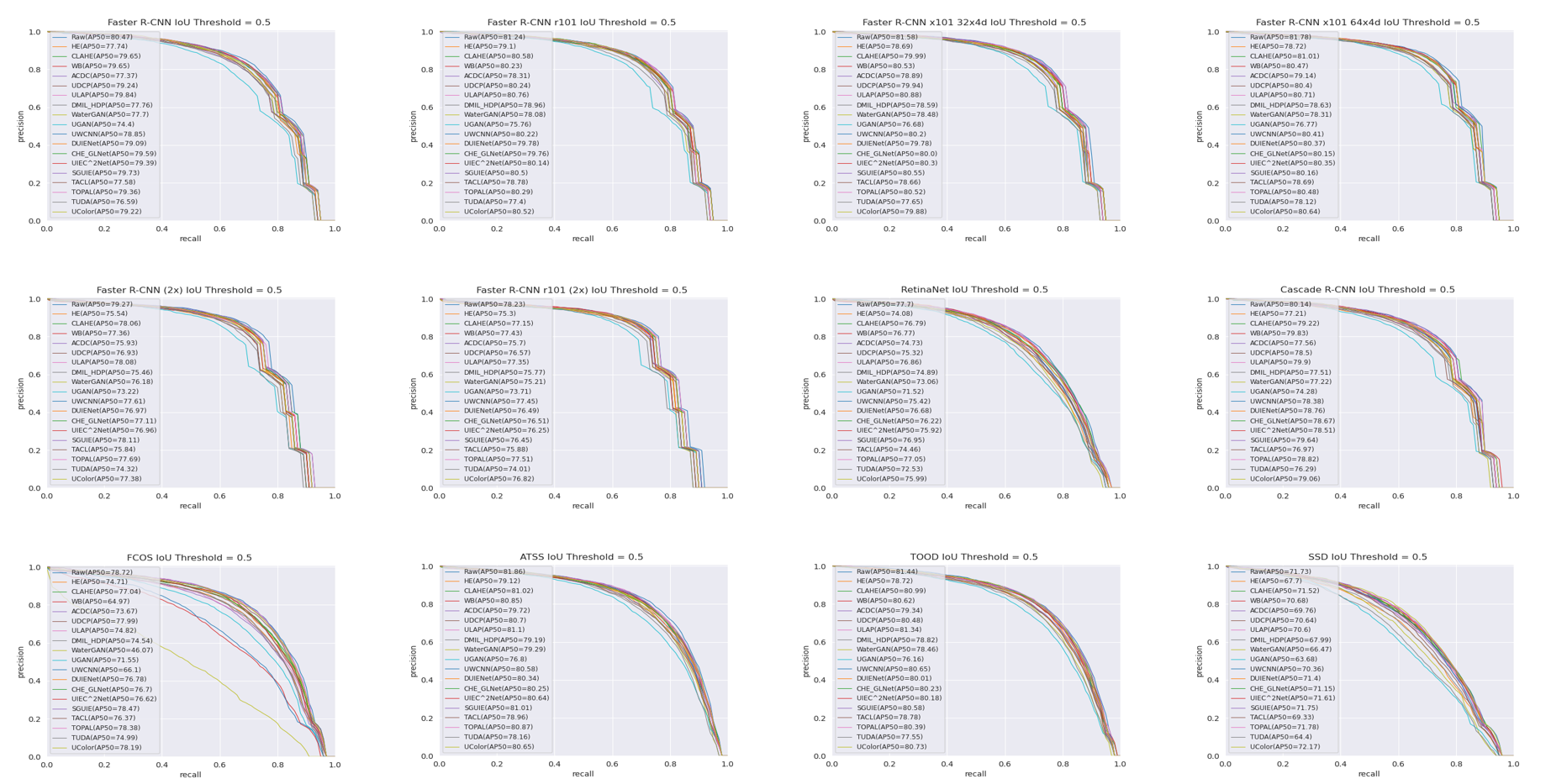}
  \caption{\textbf{Precision-recall curve at IoU threshold=0.5.} The PR curves (IoU threshold=0.5) of Faster R-CNN \cite{ren2015faster} with different backbones and training schedules, RetinaNet \cite{lin2017focal}, Cascade R-CNN \cite{cai2018cascade}, FCOS \cite{tian2019fcos}, ATSS \cite{zhang2020bridging}, TOOD\cite{feng2021tood}, and SSD \cite{liu2016ssd} with average PR curve of 4 categories.}
  \label{fig5}
\end{figure*}

\begin{figure*}[!t]    
  \centering
  \includegraphics[width=1.0\linewidth]{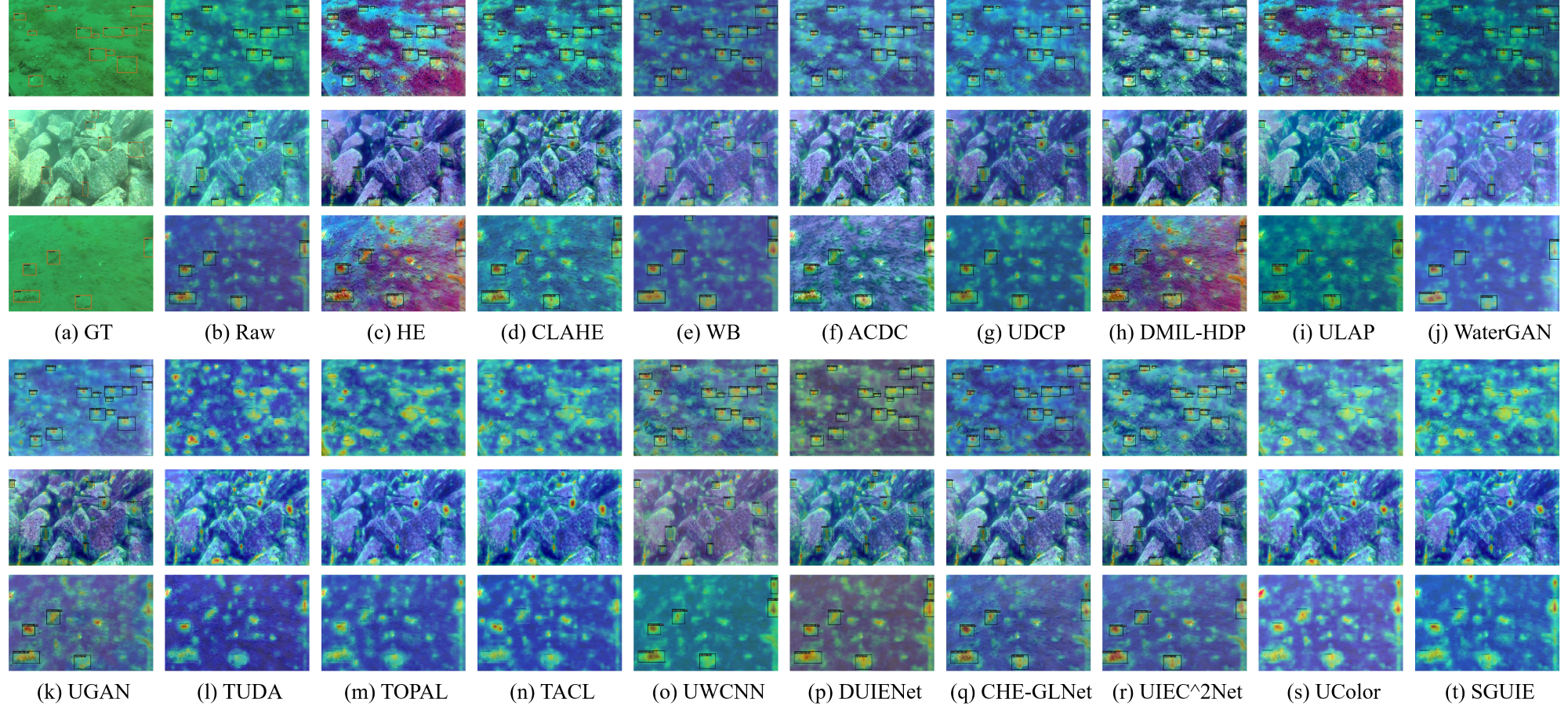}
  \caption{\textbf{Class Activation Map (CAM) visualizations on URPC2020 \cite{liu2021dataset}.} From top to bottom are the samples of the output feature map of the third block of ResNet-50 in Faster R-CNN \cite{ren2015faster}, RetinaNet \cite{lin2017focal}, and TOOD\cite{feng2021tood} that trained and tested on  (b) raw and results enhanced using (c) HE \cite{hummel1977image}, (d) CLAHE \cite{zuiderveld1994contrast}, (e) WB \cite{ebner2007color}, (f) ACDC \cite{zhang2022underwater}, (g) UDCP \cite{drews2016underwater}, (h) DMIL-HDP \cite{li2016mlhp}, (i) ULAP \cite{song2018rapid}, (j) WaterGAN \cite{li2017watergan}, (k) UGAN \cite{fabbri2018enhancing}, (l) TUDA \cite{wang2023domain}, (m) TOPAL \cite{jiang2022target}, (n) TACL \cite{liu2022twin}, (o) UWCNN \cite{li2020underwater}, (p) DUIENet \cite{li2019underwater}, (q) CHE-GLNet \cite{fu2020underwater}, (r) UIEC\^{}2Net \cite{wang2021uiec}, (s) UColor \cite{li2021underwater}, and (t) SGUIE \cite{qi2022sguie} and map it back to the input image. The color of the area from the most to least interest gradually changes from red to blue.}
  \label{fig6}
\end{figure*}

The distribution of error rates varies among different detectors.
In the underwater object detection dataset, the underwater image enhancement may decrease  the localization error and classification error in two-stage detectors but may increase both two errors in one-stage detectors. 
In general, the background error and miss GT error have the most significant impact on object detection performance.
We believe this is related to the complex background of the underwater object detection dataset we used in the  experiment. 
The categories such as holothurian, echinus, and scallop are similar to the seabed or submarine rocks. 
Since the enhancement may introduce additional noise and interference, making the background similar to the detected categories, it will bring more interference. 
For instance, the seabed will generally become black and white after underwater image enhancement, while holothurian, echinus, and scallop in the detection dataset are also white or black. 
This interference will increase the background and miss GT error, thereby reducing the overall detection performance. 
Therefore, \textit{in order to improve the performance of underwater object detection, the design of underwater image enhancement algorithms especially on how to reduce the interference caused by the background should be taken into account.}

\noindent
\textbf{Qualitative Analysis.} 
Samples of the visual detection results are shown in Fig. \ref{fig4}. As we can see in Fig. \ref{fig4}(b), the inference results on the raw images are almost identical to the GT boxes.
The UDCP \cite{drews2016underwater} domain detectors also detect all objects.
While other enhanced domain detectors more or less miss objects or incorrectly detect the background as a category.
As shown in the first row of Fig. \ref{fig4}, except for the raw and UDCP domains, other enhanced domain detectors cannot detect echinus at the bottom of the image, and the CLAHE \cite{zuiderveld1994contrast}, TUDA \cite{wang2023domain}, and SGUIE \cite{qi2022sguie} domain detector incorrectly detects background as starfish.
Similar to FCOS \cite{tian2019fcos} (the fourth row of Fig \ref{fig4}), the HE \cite{zuiderveld1994contrast}, ACDC\cite{zhang2022underwater}, ULAP \cite{song2018rapid}, WaterGAN \cite{li2017watergan}, UGAN \cite{fabbri2018enhancing}, TOPAL \cite{jiang2022target}, TACL \cite{liu2022twin}, UWCNN \cite{li2020underwater}, UColor \cite{li2021underwater}, and SGUIE \cite{qi2022sguie} domain detectors cannot detect all objects, especially, TUDA \cite{wang2023domain} domain detector does not detect anything in the image. Moreover, the WB \cite{ebner2007color}, WaterGAN \cite{li2017watergan}, TOPAL \cite{jiang2022target}, and SGUIE \cite{qi2022sguie} detect background as a category.
In the fifth row, there are three scallops in the middle right of the image. However, only the raw and UColor \cite{li2021underwater} domain detector can detect all three categories, and others have missed one or two categrories. Meanwhile, the ACDC \cite{zhang2022underwater} domain detector detects the background as an object in several places in the fifth image.
From the visual detection results samples, we can preliminary find that 
\textit{
when there is less background interface and the object is simple, all domain detectors can correctly detect the object. 
Nevertheless, when the background of the object is relatively complex, it is easy to meet the problem of missed or false detection.} 
It is noteworthy that this observation aligns with the analysis derived from TIDE.

To be more convincing, we also analyze the precision-recall (PR) curve in Fig. \ref{fig5}, in which we show the PR curves (IoU threshold=0.5) of Faster R-CNN \cite{ren2015faster} with different backbones and training schedules, RetinaNet \cite{lin2017focal}, Cascade R-CNN \cite{cai2018cascade}, FCOS \cite{tian2019fcos}, ATSS \cite{zhang2020bridging}, TOOD\cite{feng2021tood}, and SSD \cite{liu2016ssd}. A more detailed PR curves with IoU threshold of 0.75 and PR curves for each category at IoU threshold of 0.5 are shown in Appendix \ref{appendix_B}
For the PR curves, we have two important observations: 
1) The high-precision part contains high-confident detection results, and the curves are highly overlapped. That is, when the confidence score is high, the performance of each detector is almost the same.
For example, when the recall rate is less than 0.4 for the Faster R-CNN \cite{ren2015faster} with different backbone, the raw domain and enhanced domain detectors have similar performance. (Similar situation when the recall is less than 0.2, 0.3, and 0.4 in RetinaNet \cite{lin2017focal}, ATSS \cite{zhang2020bridging}, and TOOD \cite{feng2021tood}, respectively).
2) As shown in Fig. \ref{fig5}, the difference in the final recalls is not obvious, and the difference in the precision values is large when the recall is between 0.6 and 0.9. 
For instance, the precision of Faster R-CNN \cite{ren2015faster} is significantly higher than other detectors when the recall is between 0.6 and 0.8, a similar situation also exists in the recall between 0.7 to 0.9 on ATSS \cite{zhang2020bridging}.
This indicates that the impact of underwater image enhancement is mainly concentrated in the inference bounding boxes with medium to low confidence scores, which disappears in the bounding boxes with very low confidence scores.\\
Therefore, we preliminarily conclude \textit{the enhancement will inhibit the detector to detect the hard samples of the underwater object detection dataset.}

Furthermore, we also visualize some Class Activation Maps (CAM) \cite{zhou2016learning} in Fig. \ref{fig6}.
We select the output feature map of the third block of ResNet-50 \cite{he2016deep} and map it back to the input image. The color of the area from the most to least interest changes from red to blue.
As presented, the region of the interest extracted by each domain detector is essentially same. We speculate that the factors responsible for the different results (some objects cannot be detected in some detectors) are mainly small differences in the activation scores. 
For example, each domain detector has a different activation score for the upper right corner of the image in the third row, resulting in some detectors incorrectly detecting the background as an object.
Another example is that the activation score distribution of the upper region of the image in the second row obtained from UGAN \cite{fabbri2018enhancing} domain detector is significantly different from other domain detectors, which causes the UGAN domain detector fails to detect the object in that area. 
This is most likely because the enhancement destroys the original features of the object.
%
Therefore, we have further validated the aforementioned findings, providing evidence that \textit{for easy cases, all domain detectors can easily detect them, while for hard cases, the interference caused by the enhancement will cause fluctuations in feature extraction and degrade the detector performance.}

\noindent
\textbf{Study of the Relationship Between Image Enhancement and Detection.} Together with the previous analysis, we find that except for the raw domain detector, the CLAHE \cite{zuiderveld1994contrast}, ULAP \cite{song2018rapid}, and SGUIE \cite{qi2022sguie} domain detectors also obtain good performance, but the qualitative results of CLAHE and ULAP are not good as they do not recover the underwater color cast.
Meanwhile, the HE \cite{hummel1977image}, DMIL-HDP \cite{li2016mlhp}, and TACL \cite{liu2022twin} domain images achieve higher quantitative scores but obtain poor performance on their detectors. 
The result show that \textit{1) color cast is not the core interference that affects the object detection; 2) the enhancement based on good visual perception of human vision may not be beneficial for the detector; and 3) the current objective underwater image enhancement evaluation metrics cannot reflect the performance of subsequent object detection.}

Additionally, the UGAN \cite{fabbri2018enhancing}, WaterGAN \cite{li2017watergan}, and TUDA \cite{wang2023domain} domain detectors have the worst performance. As shown in Fig. \ref{fig2} and \ref{fig5}, the images processed by WaterGAN introduce noise and convert the greenish domain to multiple color domains, such as bluish, yellowish, and purplish.
Moreover, the enhancement results of UGAN \cite{fabbri2018enhancing} and TUDA \cite{wang2023domain} suffer from unclear edges and blur. Moreover, UGAN \cite{fabbri2018enhancing} also introduces color blocks.
The images processed by UWCNN \cite{li2020underwater} and UColor \cite{li2021underwater} also suffer from the problem of converting the greenish domain to yellowish or reddish domain. 
In contrast, the performance of other CNN-based enhancement (DUIENet \cite{li2019underwater}, CHE-GLNet \cite{fu2020underwater}, UIEC\textasciicircum{}2Net \cite{wang2021uiec}) domain detectors is not much different from the raw domain detector, which may be related to the fact that these enhancement algorithms do not introduce other color casts. 
The processing of HE \cite{hummel1977image} and DMIL-HDP \cite{li2016mlhp} will also introduce noise, especially the red noise, resulting in a worse performance on their domain detectors. Compared with HE \cite{hummel1977image} and DMIL-HDP \cite{li2016mlhp}, WB \cite{ebner2007color} and ACDC \cite{zhang2022underwater} can successfully remove the color cast, but WB \cite{ebner2007color} brings the low contrast problem and ACDC \cite{zhang2022underwater} brings the low saturation problem.
However, the WB domain detector achieves better performance than ACDC domain detector. We can infer from this that \textit{
1) edges can seriously affect the detector performance, therefore the enhancement algorithms need to preserve the edge information of the images considerably; 
2) color cast introduced by the underwater image enhancement may cause discontinuities in the domain, which degrades the detector performance; 
3) noise can degrade the detector performance, which suggests that we should avoid introducing extra noise when enhancing underwater images; and
4) contrast has little effect on object detection, but the color richness and saturation affect the detector.}

\section{Discussion and Conclusion}

In this paper, we conduct an empirical study to investigate the effect of underwater image enhancement on underwater object detection.
We select 18 classic and recent underwater image enhancement algorithms to pre-process the underwater object detection dataset and apply the enhanced data to retrain 7 deep learning-based detectors. 
Through the above experiments, several interesting observations and insights are obtained: 
\begin{itemize}
\item One of the most significant findings is that underwater image enhancement suppresses the performance of object detection. Especially, it inhibits the detector to detect the hard cases as the image enhancement may increase the interference caused by the background.
\item  By changing the detector backbones and training schedules, it is further found that improving the network feature extraction capability cannot reduce the negative impact of the underwater image enhancement on underwater object detection. 
\item  Although underwater image enhancement can solve the degradation problems of underwater images and obtain the images with better visual perception, it introduces other quality degradation problems that have effects on underwater object detection. 
\item  Underwater color cast is not the core interference that affects the object detection, but multiple colors introduced by the enhancement affect the performance of detectors.
\item  Underwater image enhancement may introduce noise interference, edge blur, and texture corruption problems, which seriously hurt the performance of detectors.
\item  The over-processing of image properties (contrast, saturation, and color richness) can also lead to the drops of detector performance, where saturation and color issues are the main factors that affect the detector's performance. 
\item  The experiments indicate the limitations of the existing underwater image quality assessment metrics, which not only have a gap with human visual perception but also cannot directly represent the performance of subsequent high-level tasks.
\end{itemize}

These findings can be extended to the object detection of hazy images, motion-blurred images, low-resolution images, and even high-quality images. Hence, we provide an outlook on the future research directions:  

\textit{For image enhancement}, the enhancement algorithms should have better generalization ability to avoid introducing other color cast and noise and should purposefully implement image enhancement, such as reducing the interference caused by the background.

\textit{For enhancement evaluation metrics}, it should be designed for not only considering the visual characteristics of the human vision, but also meeting the requirement of the image features extracted by machine.

\textit{For low-quality image object detection}, it should be designed to improve both the performance of detectors and combine the advantages of image enhancement, such as a joint application of image enhancement and object detection.
We will continue to explore on how to design task-oriented degraded image enhancement and evaluation metrics, and the joint application of degraded image enhancement and object detection in the future. 

\newpage

\bibliographystyle{IEEEtran}
\bibliography{mybibfile}

\clearpage
\newpage

\onecolumn

\begin{center}
    \Large\textbf{{Appendix}}\\
\end{center}

\titleformat{\section}[hang]{\normalfont\large\bfseries}{\thesection}{1em}{\raggedright}

\section*{A. Experimental Results on RUOD Dataset}
\label{appendix_A}

\begin{figure}[h]     
  \centering
  \includegraphics[width=0.55\linewidth]{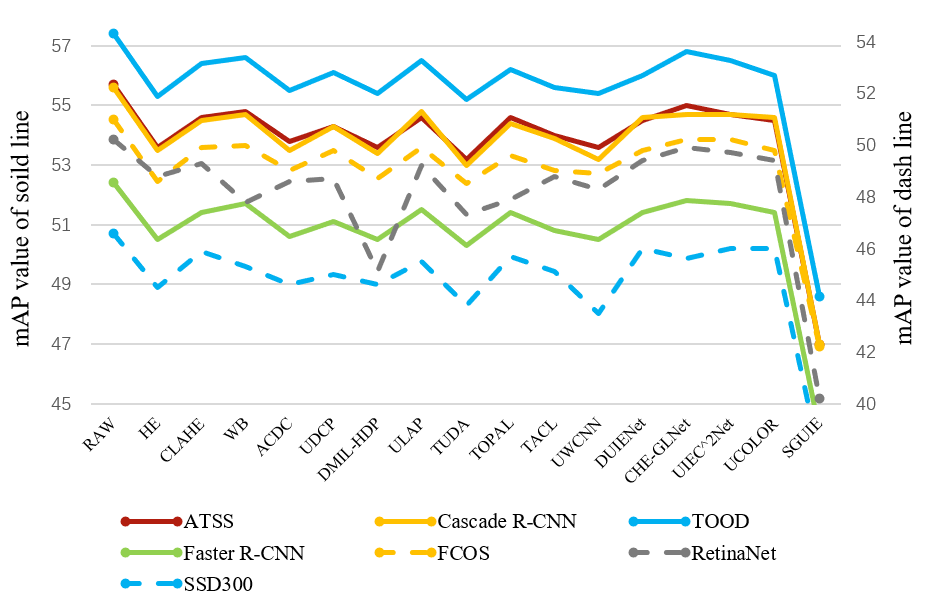}
  \caption{\textbf{The line graph of AP values on RUOD dataset.} The line graph of AP values of Faster R-CNN \cite{ren2015faster}, Cascade R-CNN \cite{cai2018cascade}, RetinaNet \cite{lin2017focal}, FCOS \cite{tian2019fcos}, ATSS \cite{zhang2020bridging}, TOOD \cite{feng2021tood}, and SSD \cite{liu2016ssd} on RUOD \cite{fu2023rethinking} dataset.}
  \label{fig7}
\end{figure}

We retrain the detectors separately and use the same corresponding images (the same domain) for testing. Experiment settings are same as URPC2020 \cite{liu2021dataset}, which can be found in Section \ref{exp_setting}.
The results of different detectors with different domains on RUOD \cite{fu2023rethinking} dataset are shown in Table \ref{table6}, and the corresponding line graphs are shown in Fig. \ref{fig7}.

\begin{table*}[h]    
	\centering
	\caption{\textbf{The AP result (\%) of Faster R-CNN \cite{ren2015faster}, Cascade R-CNN \cite{cai2018cascade}, RetinaNet \cite{lin2017focal}, FCOS \cite{tian2019fcos}, ATSS \cite{zhang2020bridging}, TOOD \cite{feng2021tood}, and SSD \cite{liu2016ssd} that training with raw and different enhanced RUOD dataset \cite{fu2023rethinking} and testing with the same domain.} The highest performance in \textcolor{rred}{red}, the second best value is in \textcolor{bblue}{blue}, and the third best value is in \textcolor{yyellow}{yellow}. We add subscripts after the AP values, which represent the current ranking of the detector performance.}
	\resizebox{\linewidth}{!}{
	\begin{tabular}{c|ccccccc|c}
			\hline
\textbf{Methods}           & \textbf{\makecell{  Faster  \\   R-CNN\cite{ren2015faster}}} & \textbf{\makecell{  Cascade   \\  R-CNN  \cite{cai2018cascade}}} & \textbf{RetinaNet \cite{lin2017focal}} & \textbf{FCOS \cite{tian2019fcos}} & \textbf{ATSS \cite{zhang2020bridging}} & \textbf{TOOD \cite{feng2021tood}} & \textbf{SSD \cite{liu2016ssd}} & \textbf{\makecell{Average   \\  Ranking}} \\ \hline
RAW                                              &  \cellcolor{rred}{$52.4_{1}$}  &	\cellcolor{rred}{$55.6_{1}$}  &	\cellcolor{rred}{$50.2_{1}$}  &	\cellcolor{rred}{$51.0_{1}$}  &	\cellcolor{rred}{$55.7_{1}$}  &	\cellcolor{rred}{$57.4_{1}$}  &	\cellcolor{rred}{$46.6_{1}$}  &	\cellcolor{rred}{1.00} 	\\
HE \cite{hummel1977image}                        &  {$50.5_{13}$}  	&	{$53.5_{12}$}  	&	{$48.8_{8}$}  &	{$48.6_{15}$}  	&	{$53.6_{13}$}  	   &	{$55.3_{15}$}  &	{$44.5_{14}$}  	&	12.86 	\\
CLAHE \cite{zuiderveld1994contrast}              &  {$51.4_{6}$}  	&	{$54.5_{8}$}  	&	{$49.3_{6}$}  &	{$49.9_{5}$}  	&	{$54.6_{5}$}  	   &	{$56.4_{6}$}   &	{$45.9_{5}$}  	&	5.86 	\\
WB \cite{ebner2007color}                         &  \cellcolor{yyellow}{$51.7_{3}$}  &	\cellcolor{yyellow}{$54.7_{3}$}  &	{$47.8_{14}$}  &	{$50.0_{4}$}  &	\cellcolor{yyellow}{$54.8_{3}$}  &	\cellcolor{yyellow}{$56.6_{3}$}  &	{$45.3_{9}$}  &	5.57 	\\
ACDC \cite{zhang2022underwater}                  &  {$50.6_{12}$}  	&	{$53.5_{12}$}  	&	{$48.6_{11}$}  &	{$49.0_{11}$}  &	{$53.8_{12}$}  &	{$55.5_{12}$}  &	{$44.6_{12}$}  &	11.71 	\\
UDCP \cite{drews2016underwater}                  &  {$51.1_{10}$}  	&	{$54.3_{10}$}  	&	{$48.7_{10}$}  &	{$49.8_{7}$}   &	{$54.3_{10}$}  &	{$56.1_{8}$}   &	{$45.0_{11}$}  &	9.43 	\\
DMIL-HDP \cite{li2016mlhp}                       &  {$50.5_{13}$}  	&	{$53.4_{14}$}  	&	{$45.1_{16}$}  &	{$48.7_{14}$}  &	{$53.6_{13}$}  &	{$55.4_{13}$}  &	{$44.6_{12}$}  &	13.57 	\\
ULAP \cite{song2018rapid}                        &  {$51.5_{5}$}  	&	\cellcolor{bblue}{$54.8_{2}$}  &	{$49.2_{7}$}  &	{$49.9_{5}$}  &	{$54.6_{5}$}  &	{$56.5_{4}$}  &	{$45.5_{8}$}  &	5.14 	\\
TUDA \cite{wang2023domain}                       &  {$50.3_{16}$} 	&	{$53.0_{16}$}  	&	{$47.3_{15}$}  &	{$48.5_{16}$}  &	{$53.2_{16}$}  &	{$55.2_{16}$}  &	{$43.8_{15}$}  &	15.71 	\\
TOPAL \cite{jiang2022target}                     &  {$51.4_{6}$}  	&	{$54.4_{9}$}  	&	{$47.9_{13}$}  &	{$49.6_{10}$}  &	{$54.6_{5}$}   &	{$56.2_{7}$}   &	{$45.7_{6}$}  &	8.00 	\\
TACL \cite{liu2022twin}                          &  {$50.8_{11}$}  	&	{$53.9_{11}$} 	&	{$48.8_{8}$}   &	{$49.0_{11}$}  &	{$54.0_{11}$}  &	{$55.6_{11}$}  &	{$45.1_{10}$}  &	10.43 	\\
UWCNN \cite{li2020underwater}                       &  {$50.5_{13}$}  	&	{$53.2_{15}$} 	&	{$48.3_{12}$}  &	{$48.9_{13}$}  &	{$53.6_{13}$}  &	{$55.4_{13}$}  &	{$43.5_{16}$}  &	13.57 	\\
DUIENet \cite{li2019underwater}                  &  {$51.4_{6}$}  	&	{$54.6_{6}$}  	&	{$49.4_{4}$}   &	{$49.8_{7}$}   &	{$54.5_{8}$}   &	{$56.0_{9}$}   &	\cellcolor{bblue}{$46.0_{2}$}  &	6.00 	\\
CHE-GLNet \cite{fu2020underwater}                &  \cellcolor{bblue}{$51.8_{2}$}  	&	\cellcolor{yyellow}{$54.7_{3}$}  &	\cellcolor{bblue}{$49.9_{2}$}  &	\cellcolor{bblue}{$50.2_{2}$}  &	\cellcolor{bblue}{$55_{2}$}  &	\cellcolor{bblue}{$56.8_{2}$}  &	{$45.6_{7}$}  &	\cellcolor{bblue}{2.86} 	\\
UIEC\textasciicircum{}2Net \cite{wang2021uiec}   &  \cellcolor{yyellow}{$51.7_{3}$}  &	\cellcolor{yyellow}{$54.7_{3}$}  &	\cellcolor{yyellow}{$49.7_{3}$}  &	\cellcolor{bblue}{$50.2_{2}$}  &	{$54.7_{4}$}  &	{$56.5_{4}$}  &	\cellcolor{bblue}{$46.0_{2}$}  &	\cellcolor{yyellow}{3.00} 	\\
UColor \cite{li2021underwater}                   &  {$51.4_{6}$}  	&	{$54.6_{6}$}  	&	{$49.4_{4}$}   &	{$49.8_{7}$}   &	{$54.5_{8}$}   &	{$56.0_{9}$}   &	\cellcolor{bblue}{$46.0_{2}$}  &	6.00 	\\
SGUIE \cite{qi2022sguie}                         &  {$44.1_{17}$}   &	{$47.0_{17}$}  	&	{$40.2_{17}$}  &	{$42.2_{17}$}  &	{$47.0_{17}$}  &	{$48.6_{17}$}  &	{$38.0_{17}$}  &	17.00 	\\
\hline
	\end{tabular}}
	\label{table6}
\end{table*}

The experimental results are similar to the findings in Section \ref{exp_det}, in particular, all detectors retrained on the raw domain can achieve the highest AP values than the detectors retrained on other domains. 
In addition, all detectors follow a similar trend, and TOOD \cite{feng2021tood} achieves the best results, followed by ATSS \cite{zhang2020bridging} and Cascade R-CNN \cite{cai2018cascade}, 
Except for SGUIE \cite{qi2022sguie}, the rank of different enhancement domains is similar to URPC2020 experimental results.
The CNN-based enhancement domain detectors, including CHE-GLNet \cite{fu2020underwater}, UIEC\^{}2Net \cite{wang2021uiec}, DUIENet \cite{li2019underwater}, and UColor \cite{li2021underwater} achieve better performance, CLAHE \cite{zuiderveld1994contrast}, WB \cite{ebner2007color}, and ULAP \cite{song2018rapid} can also achieve high AP values.
In contrast, the performance of the detectors retrained on the HE \cite{hummel1977image}, DMIL-HDP \cite{li2016mlhp}, TUDA \cite{wang2023domain}, and UWCNN \cite{li2020underwater} domains degrade severely.
Therefore, we also conclude that \textit{underwater image enhancement suppresses the performance of object detection} on RUOD \cite{fu2023rethinking} dataset.

\section*{B. Supplementary Analysis of PR Curves on URPC2020} \label{appendix_B}

\begin{figure*}[h]    
  \centering
  \includegraphics[width=0.9\linewidth]{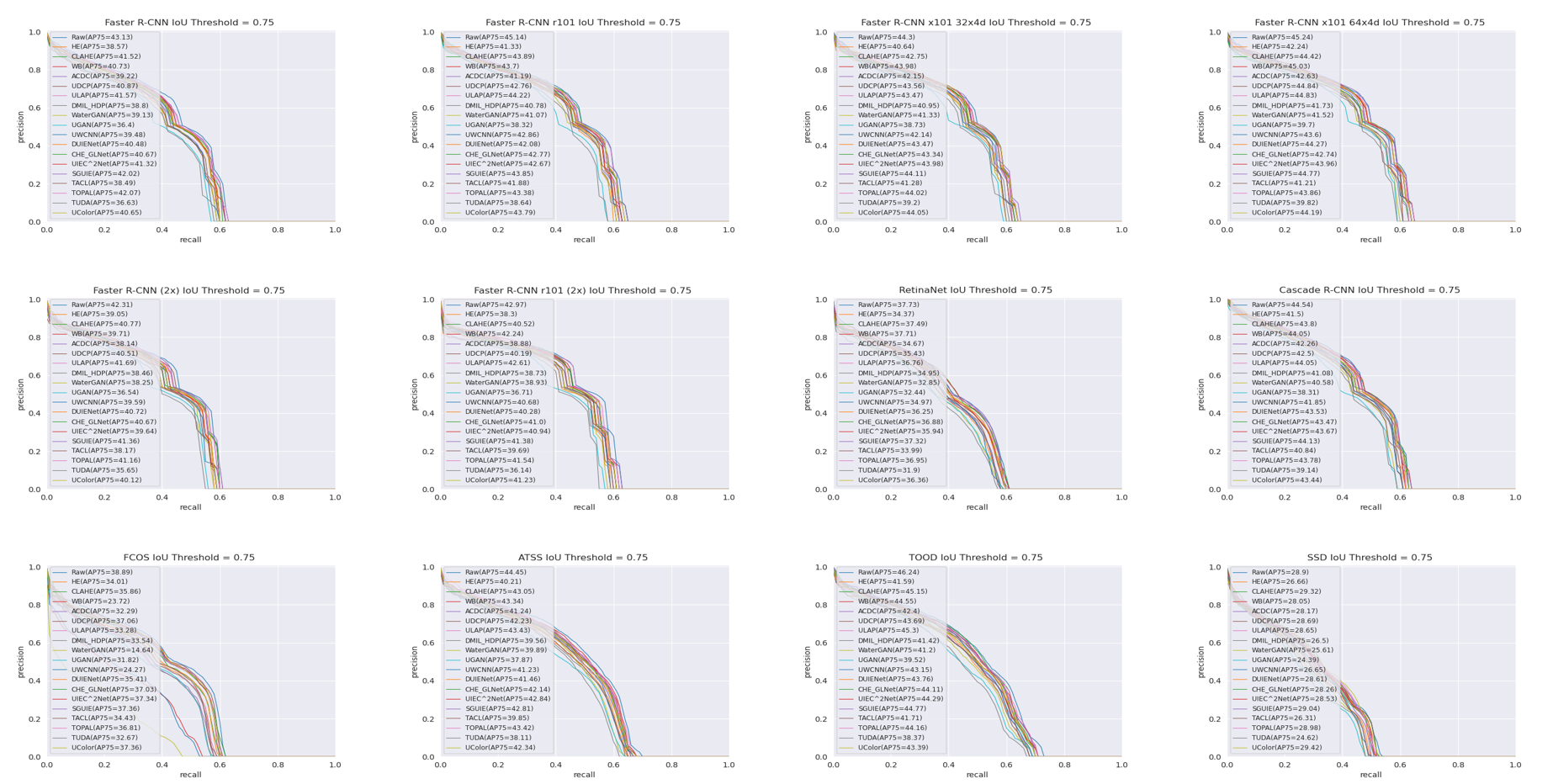}
  \caption{\textbf{Precision-recall curve at IoU threshold=0.75.} The PR curves (IoU threshold=0.75) of Faster R-CNN \cite{ren2015faster} with different backbones and training schedules, RetinaNet \cite{lin2017focal}, Cascade R-CNN \cite{cai2018cascade}, FCOS \cite{tian2019fcos}, ATSS \cite{zhang2020bridging}, TOOD\cite{feng2021tood}, and SSD \cite{liu2016ssd} with average PR curve of 4 categories on URPC2020 dataset.}
  \label{fig8}
\end{figure*}

To further analyze "How does underwater image enhancement contribute to underwater object detection?", we also show the PR curves of Faster R-CNN \cite{ren2015faster} with different backbones and training schedules, RetinaNet \cite{lin2017focal}, Cascade R-CNN \cite{cai2018cascade}, FCOS \cite{tian2019fcos}, ATSS \cite{zhang2020bridging}, TOOD\cite{feng2021tood}, and SSD \cite{liu2016ssd} with IoU threshold=0.75 in Fig. \ref{fig8} and 4 categories (echinus, holothurian, scallop, and starfish) with IoU threshold=0.5 in Fig. \ref{fig9}.

\begin{figure*}[h]    
  \centering
  \includegraphics[width=0.9\linewidth]{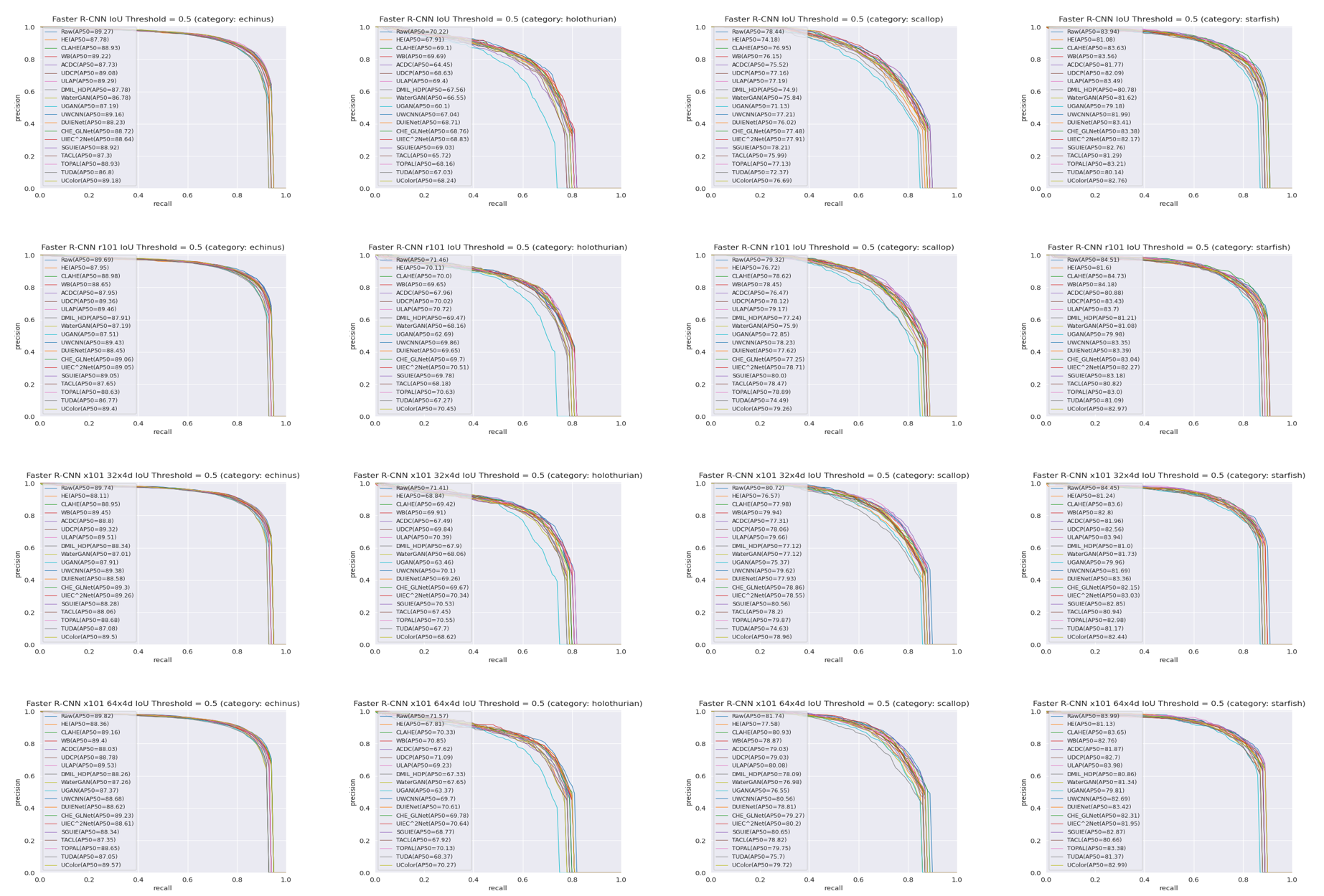}
\end{figure*}

\begin{figure*}[h]    
  \centering
  \includegraphics[width=0.9\linewidth]{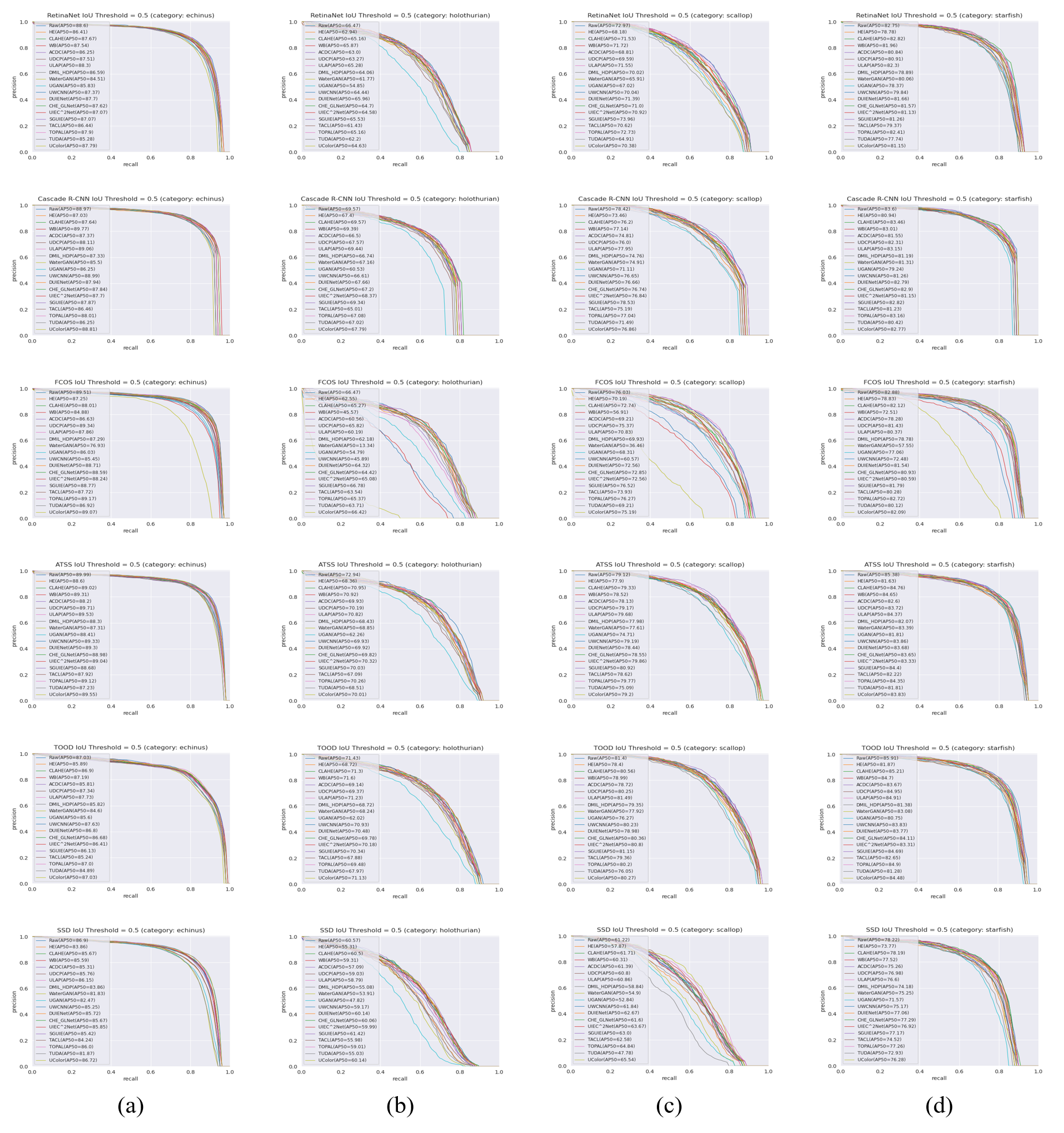}
  \caption{\textbf{Precision-recall curve at IoU threshold=0.5.} From top to bottom are the PR curve of Faster R-CNN \cite{ren2015faster}, RetinaNet \cite{lin2017focal}, Cascade R-CNN \cite{cai2018cascade}, FCOS \cite{tian2019fcos}, ATSS \cite{zhang2020bridging}, TOOD\cite{feng2021tood}, and SSD \cite{liu2016ssd} with (a) PR curve of echinus, (b) PR curve of holothurian, (c) PR curve of scallop, and (d) PR curve of starfish. (Zoom in for the best view.)}
  \label{fig9}
\end{figure*}

According to Fig. \ref{fig8}, we further observe that compare to IoU threshold=0.5, the difference in the final recalls is obvious, and the curves diverge at the very beginning. 
This indicates that as the localization requirements become more stringent during computing True Positive (TP), the performance of different domain detectors becomes significantly different.
For example, the precision of Faster R-CNN \cite{ren2015faster} is significantly higher than other detectors when the recall is between 0.2 and 0.6, a similar situation also exists on ATSS\cite{zhang2020bridging}. 
In addition, as shown in Fig. \ref{fig9}, the observations are similar to Fig. \ref{fig5}. \textit{i.e.}, The high-precision part contains high-confident detection results, and the curves are highly overlapped. That is, when the confidence score is high, the performance of each detector is almost the same. For example, when the recall rate is less than 0.6 for the "echinus" category, the raw domain and enhanced domain detectors have similar performance. (Similar situation when the recall is less than 0.2 and 0.6 for scallop's and starfish's PR curve, respectively). 
We also observed that the divergence of PR curves varies across different categories. 
For instance, the PR curves of "echinus" exhibit a higher degree of overlap, while "holothurian" show more significant discrepancies, indicating variations in recognition difficulty across different categories.

\end{document}